\documentclass[10pt, twoside]{IEEEtran} \twocolumn

\usepackage{caption, subcaption, graphicx, epstopdf, algorithm,amsmath,amssymb, url,algorithmic,epsfig}
\usepackage{multirow}
\newcommand{\beq}{\begin{equation}}
\newcommand{\eeq}{\end{equation}}
\newcommand{\bdm}{\begin{displaymath}}
\newcommand{\edm}{\end{displaymath}}

\newtheorem{theorem}{Theorem}

\newtheorem{definition}[theorem]{Definition}

\newcommand{\bd}{\begin{definition}}
\newcommand{\ed}{\end{definition}}

\newcommand{\bv}{\begin{vugraph}}
\newcommand{\ev}{\end{vugraph}}
\newcommand{\bi}{\begin{itemize}}
\newcommand{\ei}{\end{itemize}}
\newcommand{\ben}{\begin{enumerate}}
\newcommand{\een}{\end{enumerate}}

\newcommand{\bean}{\begin{eqnarray*} }
\newcommand{\eean}{\end{eqnarray*} }
\newcommand{\bea}{\begin{eqnarray} }
\newcommand{\eea}{\end{eqnarray} }

\newcommand{\ba}{\begin{array} }
\newcommand{\ea}{\end{array} }

\newcommand{\new}{\text{new}}

\usepackage{bm}

\title{Video Denoising and Enhancement via\\ Dynamic Video Layering}

\author{Han Guo and Namrata Vaswani  \\ Iowa State University, Ames, IA, USA \\  Email: \{hanguo,namrata\}@iastate.edu \thanks{This work was partially supported by Rockwell Collins. A portion of this work was presented at IEEE Workshop on Statistical Signal Processing (SSP) 2016 \cite{guo2016video}. }}

\date{}

\begin{document}
%
\maketitle
\begin{abstract}
Video denoising refers to the problem of removing ``noise" from a video sequence. Here the term ``noise" is used in a broad sense to refer to any corruption or outlier or interference that is not the quantity of interest. In this work, we develop a novel approach to video denoising that is based on the idea that many noisy or corrupted videos can be split into three parts - the ``low-rank layer",  the ``sparse layer", and  a small residual  (which is small and bounded). 
We show, using extensive experiments, that our denoising approach outperforms the state-of-the-art denoising algorithms.
\end{abstract}

\section{Introduction}
Video denoising refers to the problem of removing ``noise" from a video sequence. Here the term ``noise" is used in a broad sense to refer to any corruption or outlier or interference that is not the quantity of interest.

In the last few decades there has been a lot of work on video denoising. Many of the approaches extend image denoising ideas to 3D by also exploiting dependencies across the temporal dimension.  An important example of this is  ``grouping and collaborative filtering'' approaches which try to search for similar image patches both within an image frame and across nearby frames, followed by collaboratively filtering the noise from the stack of matched patches \cite{buades2005image, dabov2007image, elad2006image,foi2007pointwise,mairal2009non}.
One of the most effective methods for image denoising, Block Matching and 3D filtering (BM3D) \cite{dabov2007image}, is from this category of techniques. In BM3D, similar image blocks are stacked in a 3D array followed by applying a noise shrinkage operator in a transform domain. In its video version,  VBM3D \cite{dabov2007video}, the method is generalized to video denoising by searching for similar blocks across multiple frames.
Other related works \cite{ji2010robust,ji2011robust} apply batch matrix completion or matrix decomposition on grouped image patches to remove outliers.

Other recent works on video denoising include approaches that use motion compensation algorithms from the video compression literature followed by denoising of similar nearby blocks \cite{liu2010high, guo2007temporal}; and approaches that use wavelet transform based  \cite{rahman2007video,rabbani2012video,yu2010video} and discrete cosine transform (DCT) based \cite{joachimiak2012multiview} denoising solutions. Very recent video denoising methods include algorithms based on learning a sparsifying transform \cite{ravishankar2013learning, ravishankar2014online1, ravishankar2014online2, wen2015video}. Within the image denoising literature, the most promising recent approaches are based on deep learning \cite{xie2012image, agostinelli2013adaptive}.


{\bf Contribution. }
In this paper, we develop a novel approach to video denoising, called {\em ReProCS-based Layering Denoising (ReLD)}, that is based on the idea that many noisy or corrupted videos can be split into three parts or layers - the ``low-rank layer", the ``sparse layer'' and the ``small bounded residual layer". Here these terms mean the following.
``Low-rank layer'' $\bm{\ell}_t$: the matrix formed by each vectorized image of this layer is low-rank.
``Sparse layer'' $\bm{s}_t$: each vectorized image of this layer is sparse.
``Small bounded residual layer": each vectorized image of this layer has $\bm{l}_\infty$ norm that is bounded. Our proposed algorithm consists of two parts. At each time instant, it first separates the video into ``noisy" versions of the two layers $\bm{\ell}_t$ and $\bm{s}_t$. This is followed by applying an existing state-of-the-art denoising algorithm, VBM3D \cite{dabov2007video} on each layer. For separating the video, we use a combination of an existing batch technique for sparse + low-rank matrix decomposition called principal components' pursuit (PCP)  \cite{rpca} and an appropriately modified version of our recently proposed dynamic robust PCA technique called Recursive Projected Compressive Sensing (ReProCS) \cite{rrpcp_perf,guo2014online}. We initialize using PCP and use ReProCS afterwards to separate the video into a ``sparse layer'' and a ``low-rank layer''.  The video-layering step is followed by VBM3D on each of the two layers. In doing this, VBM3D exploits the specific characteristics of each layer and, hence, is able to find more matched blocks to filter over, resulting in better denoising performance. The motivation for picking ReProCS for the layering task is its superior performance in earlier video experiments involving videos with large-sized sparse components and/or significantly changing background images.


The performance of our algorithm is compared with PCP \cite{rpca}, GRASTA \cite{grass_undersampled} and non-convex rpca (NCRPCA) \cite{netrapalli2014non}, which are some of the best solutions from the existing sparse + low-rank (S+LR) matrix decomposition literature 
\cite{rpca, grass_undersampled, netrapalli2014non, Torre03aframework, rpcal0sur, frpca,xu_nips2013_1,xu_nips2013_2, mateos_anomaly} 
followed by VBM3D for the denoising step; as well as with just using VBM3D directly on the video. We also compare with one neural network based image denoising method that uses a Multi Layer Perceptron (MLP)  \cite{burger2012image} and with the approach of \cite{ji2011robust} which performs standard sparse $+$ low-rank (S+LR) matrix approximation (SLMA) on grouped image patches. As we show, our approach is 6-10 times faster than SLMA while also having improved performance. The reason is we use a novel online algorithm (after a short batch initialization) for S+LR and then use VBM3D on each layer.

\subsection{Example applications}
A large number of videos that require denoising/enhancement can be accurately modeled in the above fashion. Some examples are as follows. All videos referenced below are posted at \url{http://www.ece.iastate.edu/~hanguo/denoise.html}.
\begin{enumerate}

\item\label{sp_noise} In a traditional denoising scenario, consider slowly changing videos that are corrupted by salt-and-pepper noise (or other impulsive noise). For these types of videos, the large magnitude part of the noise forms the ``sparse layer'', while the video-of-interest (slowly-changing in many applications, e.g., waterfall, waving trees, sea water moving, etc) forms the approximate ``low-rank layer''. The approximation error in the low-rank approximation forms the ``small bounded residual". See the waterfall-salt-pepper video for an example.
    The goal is to denoise or extract out the ``low-rank layer". 

\item\label{gaussian_noise} More generally, consider slow-changing videos corrupted by very large variance white Gaussian noise. As we explain below, large Gaussian noise can, with high probability, be split into a very sparse noise component plus bounded noise. Thus, our approach also works on this type of videos, and in fact, in this scenario, we show that it significantly outperforms the existing state-of-the-art video denoising approaches.  (See Fig. \ref{VisualCurtain}.)

\item In very low-light videos of moving targets/objects (the moving target is barely visible), the denoising goal is to ``see'' the barely visible moving targets (sparse). These are hard to see because they are corrupted by slowly-changing background images (well modeled as forming the low-rank layer plus the residual). The dark-room video is an example of this. The goal is to extract out the sparse targets or, at least, the regions occupied by these objects. (See Fig. \ref{dark}.)


\end{enumerate}

Moreover, in all these examples, it is valid to argue that the columns of the low-rank matrix lie in a low-dimensional subspace that is either fixed or slowly changing. This is true, for example, when the background consists of moving waters, or the background changes are due to illumination variations. These also result in global (non-sparse) changes. In special cases where foreground objects are also present, the video itself become ``low-rank + sparse''. In such a scenario, the ``sparse layer'' that is extracted out will consist of the foreground object and the large magnitude part of the noise. Some examples are the curtain and lobby videos. The proposed ReLD algorithm works for these videos if VBM3D applied to the foreground layer video is able to separate out the foreground moving object(s) from the noise. 
%


\newcommand{\T}{\mathcal{T}}
\subsection{Problem formulation}
Let $\bm{m}_t$ denote the image at time $t$ arranged as a 1D vector of length $n$. We consider denoising for videos in which each image can be split as

\[
\bm{m}_t = \bm{\ell}_t + \bm{s}_t + \bm{w}_t
\]

where $\bm{s}_t$ is a sparse vector, $\bm{\ell}_t$'s lie in a fixed or slowly changing low-dimensional subspace of $\mathbb{R}^n$ so that the matrix $\bm{L}:=[\bm{\ell}_1, \bm{\ell}_2, \dots ,\bm{\ell}_{t_\text{max}}]$ is low-rank, and $\bm{w}_t$ is the residual noise that satisfies $\|\bm{w}_t\|_\infty \le b_w$. We use $\T_t$ to denote the support set of $\bm{s}_t$, i.e., $\T_t:=\mathrm{support}(\bm{s}_t)$.

In the first example given above, the moving targets' layer is $\bm{s}_t$, the slowly-changing dark background is $\bm{\ell}_t + \bm{w}_t$. The layer of interest is $\bm{s}_t$.
In the second example, the slowly changing video is $\bm{\ell}_t + \bm{w}_t$, while the salt-and-pepper noise is $\bm{s}_t$. The layer of interest is $\bm{\ell}_t$.
In the third example, the slowly changing video is $\bm{\ell}_t + \bm{w}_{1,t}$ with $\bm{w}_{1,t}$ being the residual; and, as we explain next, with high probability (whp), white Gaussian noise can be split as $\bm{s}_t + \bm{w}_{2,t}$ with $\bm{w}_{2,t}$ being bounded. In this case, $\bm{w}_t = \bm{w}_{1,t} + \bm{w}_{2,t}$.

Let $\bm{n}$ denote a Gaussian noise vector in $\mathbb{R}^n$ with zero mean and covariance $\sigma^2 \bm{I}$. Let $\beta(b):= 2\Phi(b) - 1$ with $\Phi(z)$ being the cumulative distribution function (CDF) of the standard Gaussian distribution.
Then, it is not hard to see that $\bm{n}$ can be split as
\[
\bm{n} = \bm{s} + \bm{w}
\]
where $\bm{w}$ is bounded noise with $\|\bm{w}\|_\infty \le b_0$ and $\bm{s}$ is a sparse vector with support size $|\T_t| \approx \left(1-\beta\left(\frac{b_0}{\sigma}\right)\right)n$ whp. More precisely, with probability at least $1 - 2 \exp(-2\epsilon^2 n)$,
\[
\left(1-\beta\left(\frac{b_0}{\sigma}\right) - \epsilon \right)n  \le |\T_t| \le \left(1-\beta\left(\frac{b_0}{\sigma} \right) + \epsilon \right)n.
\]
In words, whp, $\bm{s}$ is sparse with support size roughly $(1-\beta)n$ where $\beta = \beta\left(\frac{b_0}{\sigma} \right)$.
The above claim is a direct consequence of Hoeffding's inequality
for a sum of independent Bernoulli random variables\footnote{If $p$ is the probability of $z_i = 1$, then Hoeffding's inequality says that: 
\begin{equation*}
\Pr( (p-\epsilon) n \le  \sum_i z_i \le (p+\epsilon) n    ) \ge 1 - 2 \exp(-2\varepsilon^2 n)
\end{equation*}
We apply it to the Bernoulli random variables $z_i$'s with $z_i$ defined as $z_i=1$ if $\{\bm{s}_i \neq 0\} $ and $z_i  = 0$ if $\{\bm{s}_i = 0\} $. Clearly, $\Pr(z_i = 0) = \Pr(\bm{s}_i = 0) = \Pr(\bm{n}_i ^2 \le b_0^2) = \Phi(b_0 / \sigma) - \Phi(-b_0 / \sigma) = 2 \Phi(b_0/ \sigma) - 1 = \beta(b_0 / \sigma)$.
}.

%
%
%

\section{ReProCS-based Layering Denoising (ReLD)}


We summarize the ReProCS-based Layering Denoising (ReLD) algorithm in Algorithm \ref{Alg1} and detail each step in Algorithm \ref{Alg2}. The approach is explained below.

\begin{algorithm}
\caption{\small{Overall ReLD Algorithm}}\label{Alg1}
\begin{enumerate}
\item For $t < t_0$, initialization using PCP \cite{rpca}.

\item For all $t > t_0$, implement an appropriately modified ReProCS algorithm
\begin{enumerate}
\item  Split the video frame $\bm{m}_t$ into layers  $\hat{\bm{\ell}}_t$ and $\hat{\bm{s}}_t$

\item For every $\alpha$ frames, perform subspace update, i.e., update $\hat{\bm{P}}_t$
\end{enumerate}

\item Denoise using VBM3D

\end{enumerate}
\end{algorithm}

\begin{algorithm}
\caption{\small{Details of each step of ReLD}}\label{Alg2}
\textbf{Parameters:} We used $\alpha = 20, K_{\min}=3, K_{\max}=10, t_0=50$ in all experiments.
\begin{enumerate}
\item Initialization using PCP \cite{rpca}: Compute $(\hat{\bm{L}}_0, \hat{\bm{S}}_0) \leftarrow \text{PCP}(\bm{M}_0)$ and compute $[\hat{\bm{P}}_0,\hat{\bm{\Sigma}}_0] \leftarrow \text{approx-basis}(\hat{\bm{L}}_0,90\%)$. The notation PCP$(\bm{M})$ means implementing the PCP algorithm on matrix $\bm{M}$ and $\bm{P}$ = \text{approx-basis}$(\bm{M},b\%)$ means that $\bm{P}$ is the $b\%$ left singular vectors' matrix for $\bm{M}$.\\
Set $\hat{r}\leftarrow \text{rank}(\bm{\hat{P}}_0)$, $\hat{\bm{\sigma}}_{\min}\leftarrow (\bm{\hat{\Sigma}}_0)_{\hat{r},\hat{r}}$, $\hat{t}_0 =t_0$, flag$=$detect

\item For all $t > t_0$, implement an appropriately modified ReProCS algorithm
\begin{enumerate}
\item  Split $\bm{m}_t$ into layers $\bm{\hat{\ell}}_t$ and $\bm{\hat{s}}_t$:
\begin{enumerate}
\item Compute $\bm{y}_t\leftarrow  \bm{\Phi}_t \bm{m}_t$ with $\bm{\Phi}_t \leftarrow \bm{I} - \hat{\bm{P}}_{t-1}\hat{\bm{P}}_{t-1}'$
\item Compute $\hat{\bm{s}}_t$ as the solution of 
\begin{equation*}\label{CS}
\min_{\bm{x}} \| \bm{x} \|_1 \text{s.t.} \|   \bm{y}_t -  \bm{\Phi}_t  \bm{x}  \|_2  \leq \xi
\end{equation*}
with $\xi =\| \bm{\Phi}_t \hat{\bm{\ell}}_{t-1} \|$
\item $\hat{\T}_t \leftarrow \text{Thresh}(\bm{\hat{s}_t}, \omega)$ with $\omega = 3\sqrt{\| \bm{m}_t \|^2/n}$. Here $\T\leftarrow \text{Thresh}(\bm{x},\omega)$ means that $\T = \left\lbrace i: |(\bm{x})_i |\geq \omega \right\rbrace$\\
$\hat{\bm{s}}_{t,\ast} \leftarrow \text{LS}(\bm{y}_t, \bm{\Phi}_t, \hat{\T}_t)$. Here $\hat{\bm{x}}\leftarrow \text{LS}(\bm{y},\bm{A},\T)$ means that $\hat{\bm{x}}_\T = (\bm{A}_\T'\bm{A}_\T)^{-1}\bm{A}_\T'\bm{y}$, which is least-squared estimate of $\bm{x}$ on $\T$.
\item $\hat{\bm{\ell}}_t \leftarrow \bm{m}_t - \hat{\bm{s}}_t$, $\hat{\bm{\ell}}_{t,\ast} \leftarrow \bm{m}_t - \hat{s}_{t,\ast}$
\end{enumerate}

\item Perform subspace update, i.e., update $\hat{\bm{P}}_t$ (see details in supplementary material)

\end{enumerate}

\item Denoise using VBM3D:

\begin{enumerate}
\item $\hat{\sigma}_{\text{fg}} \leftarrow \text{Std-est}([\hat{\bm{s}}_{t},\ldots, \hat{\bm{s}}_{t_0}])$\\
$\hat{\sigma}_{\text{bg}} \leftarrow \text{Std-est}([\hat{\bm{\ell}}_{t},\ldots, \hat{\bm{\ell}}_{t_0}])$. Here Std-est$(\bm{M})$ denotes estimating the standard deviation of noise from $\bm{M}$: we first subtract column-wise mean from $\bm{M}$ and then compute the standard deviation by seeing it as a vector.

\item $\hat{\bm{S}}_{\text{denoised}}\leftarrow \text{VBM3D}([\hat{\bm{s}}_{1},\ldots, \hat{\bm{s}}_{t_{\text{max}}}], \hat{\sigma}_{\text{fg}} )$\\
$\hat{\bm{L}}_{\text{denoised}}\leftarrow \text{VBM3D}([\hat{\bm{\ell}}_{1},\ldots, \hat{\bm{\ell}}_{t_{\text{max}}}],\hat{\sigma}_{\text{bg}})$. Here \text{VBM3D}$(\bm{M}, \sigma)$ implements the VBM3D algorithm on matrix $\bm{M}$ with input standard deviation~$\sigma$.

\end{enumerate}

\end{enumerate}
\textbf{Output:} $\hat{\bm{S}}$, $\hat{\bm{S}}_{\text{denoised}}$, $\hat{\bm{L}}_{\text{denoised}}$ or $\hat{\bm{I}}_{\text{denoised}}=\hat{\bm{S}}_{\text{denoised}}+\hat{\bm{L}}_{\text{denoised}}$ based on applications
\end{algorithm}

\textbf{Initialization. } 
Take $\bm{M}_0=[\bm{m}_1, \bm{m}_2,\cdots,\bm{m}_{t_{0}}]$ as training data and use PCP \cite{rpca} to separate it into a sparse matrix $[\hat{\bm{s}}_1, \hat{\bm{s}}_2,\cdots, \hat{\bm{s}}_{t_0}]$ and a low-rank matrix $[\hat{\bm{\ell}}_1, \hat{\bm{\ell}}_2,\cdots, \hat{\bm{\ell}}_{t_0}]$.  Compute the top $b\%$ left singular vectors of $[\hat{\bm{\ell}}_1, \hat{\bm{\ell}}_2,\cdots, \hat{\bm{\ell}}_{t_0}]$ and denote by $\hat{\bm{P}}_0$. Here $b\%$ left singular vectors of a matrix $\bm{M}$ refer to the left singular vectors of $\bm{M}$ whose corresponding singular values form the smallest set of singular values that contains at least $b\%$ of the total singular values' energy.

\textbf{Splitting phase. } Let $\hat{\bm{P}}_{t-1}$ be the basis matrix (matrix with orthonormal columns) for the estimated subspace of $\bm{\ell}_{t-1}$. For $t\geq t_{0} + 1$, we split $\bm{m}_t$ into $\hat{\bm{s}}_t$ and $\hat{\bm{\ell}}_t$ using prac-ReProCS \cite{guo2014online}. To do this, we first project $\bm{m}_t$ onto the subspace orthogonal to range($\hat{\bm{P}}_{t-1}$) to get the projected measurement vector,
\begin{equation}
\bm{y}_t: = (\bm{I}-\hat{\bm{P}}_{t-1} \hat{\bm{P}}_{t-1}')\bm{m}_t:=\bm{\Phi}_{t} \bm{m}_t.
\end{equation}
Observe that $\bm{y}_t$ can be expressed as
\begin{equation}
\bm{y_t}=\bm{\Phi}_t\bm{s}_t + \bm{\beta}_t \text{ where } \bm{\beta}_t:=\bm{\Phi}_t (\bm{\ell}_t + \bm{w}_t).
\end{equation}

Because of the slow subspace change assumption, the projection nullifies most of the contribution of $\bm{\ell}_t$ and hence $\bm{\beta}_t$ is small noise. The problem of recovering $\bm{s}_t$ from $\bm{y}_t$ becomes a traditional noisy sparse recovery/CS problem and one can use $\bm{\ell}_1$ minimization or any of the greedy or iterative thresholding algorithms to solve it.
We denote its solution by $\hat{\bm{s}}_t$, and obtain $\hat{\bm{\ell}}_t$ by simply subtracting $\hat{\bm{s}}_t$ from $\bm{m}_t$.

\textbf{Denoising phase. } We perform VBM3D on $\hat{\bm{S}}=[\hat{\bm{s}}_{1},\ldots, \hat{\bm{s}}_{t_{\text{max}}}]$ and $\hat{\bm{L}}=[\hat{\bm{\ell}}_{1},\ldots, \hat{\bm{\ell}}_{t_{\text{max}}}]$ and obtain the denoised data $\hat{\bm{S}}_{\text{denoised}}$ and $\hat{\bm{L}}_{\text{denoised}}$. Based on applications, we output different results. For example, in the low-light denoising case, our output is $\hat{\bm{S}}$ since the goal is to extract out the sparse targets. In traditional denoising scenarios, the output can be $\hat{\bm{L}}_{\text{denoised}}$ or $\hat{\bm{I}}_{\text{denoised}}$=$\hat{\bm{S}}_{\text{denoised}}$+$\hat{\bm{L}}_{\text{denoised}}$. This depends on whether the video contains only background or background and foreground. In practice, even for videos with only backgrounds, adding $\hat{\bm{S}}_{\text{denoised}}$ helps improve PSNR.


\textbf{Subspace Update phase (Optional). } In long videos the span of the $\bm{\ell}_t$'s will change with time. Hence one needs to update the subspace estimate $\hat{\bm{P}}_t$ every so often. This can be done efficiently using the projection-PCA algorithm from \cite{guo2014online}.

\section{Experiments}

Due to limited space, in this paper we only present a part of the experimental results. The complete presentation of experimental results are in the supplementary material. Video demos and all tables of peak signal to noise ratio (PSNR) comparison are also available at \url{http://www.ece.iastate.edu/~hanguo/denoise.html}. Code for ReLD as well as for all the following experiments is also posted here.

\subsection{Removing Salt $\&$ Pepper noise}

First we compare the denoising performance on two dataset -- Curtain and Lobby which are available at \url{http://www.ece.iastate.edu/~hanguo/denoise.html}. The algorithms being compared are ReLD, SLMA \cite{ji2011robust}, VBM3D and a neural network image denoising method, Multi Layer Perceptron (MLP) \cite{burger2012image}. The codes for algorithms being compared are downloaded from the authors' webpages. The available MLP code contains parameters that are trained solely from image patches that were corrupted with Gaussian noise with $\sigma =25$ and hence the denoising performance is best with $\sigma =25$ and deteriorates for other noise levels.
The noise being added to the original image frames are Gaussian ($\sigma=25$) plus $8\%$ salt and pepper noise. In Fig.\ref{LobPSNR} we show a plot of the frame-wise PSNRs for the lobby video, which shows that ReLD outperforms all other algorithms -- the PSNR is the highest in all image frames.

\subsection{Removing Gaussian noise}

Next, with different levels of Gaussian noise, we compare performance of our proposed denoising framework with video layering performed using either ReProCS \cite{guo2014online} (our proposed algorithm), or using the other robust PCA algorithms - PCP \cite{rpca}, NCRPCA \cite{netrapalli2014non}, and GRASTA \cite{grass_undersampled}. We call the respective algorithms ReLD, PCP-LD, NCRPCA-LD, and GRASTA-LD for short. We test all these on the Waterfall dataset (downloaded from Youtube \url{https://www.youtube.com/watch?v=UwSzu_0h7Bg}). Besides these Laying-Denoising algorithms, we also compare with VBM3D and MLP.  Since the video length was too long, SLMA code failed for this video.

The waterfall video is slow changing and hence is well modeled as being low-rank. We add i.i.d. Gaussian noise with different variances to the video. The video consists of 650 frames
of size $108 \times 108$.
The results are summarized in Table \ref{TabPSNR}. This table contains results on a smaller sized waterfall video. This is done because SLMA and PCP-LD become too slow for the full size video. Comparisons of the other algoithms on full size video are included in the supplementary material. ReLD has the best performance when the noise variance is large. We also show the time taken by each method in parantheses. As can be seen, ReProCS is slower than VBM3D and GRASTA-LD, but has significantly better performance than both. 

In Table \ref{TabPSNR}, we also provide comparisons on four more videos - fountain, escalator, curtain and lobby (same videos as in the salt and pepper noise experiment) with different levels of Gaussian noise. In Fig.\ref{VisualCurtain} we show sample visual comparisons for the curtain video with noise standard deviation $\sigma=70$. 
As can be seen, ReLD is able to recover more details of the images while other algorithms either fail or cause severe blurring.%

\begin{figure}
	\centering
		\includegraphics[width=62mm]{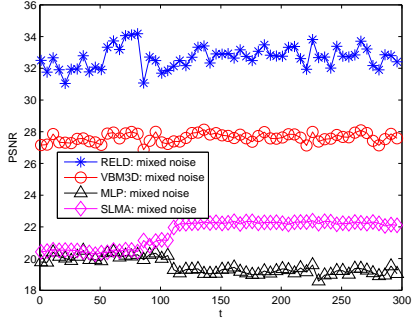}
			\vspace{-0.3cm}	
		\caption{\small{Frame-wise PSNR for Lobby dataset with Gaussian($\sigma=25$) plus salt and pepper noise.}}\label{LobPSNR}
\end{figure}

\begin{figure}[t]
	\centering
	\begin{tabular}{cc}
		\includegraphics[width=25mm]{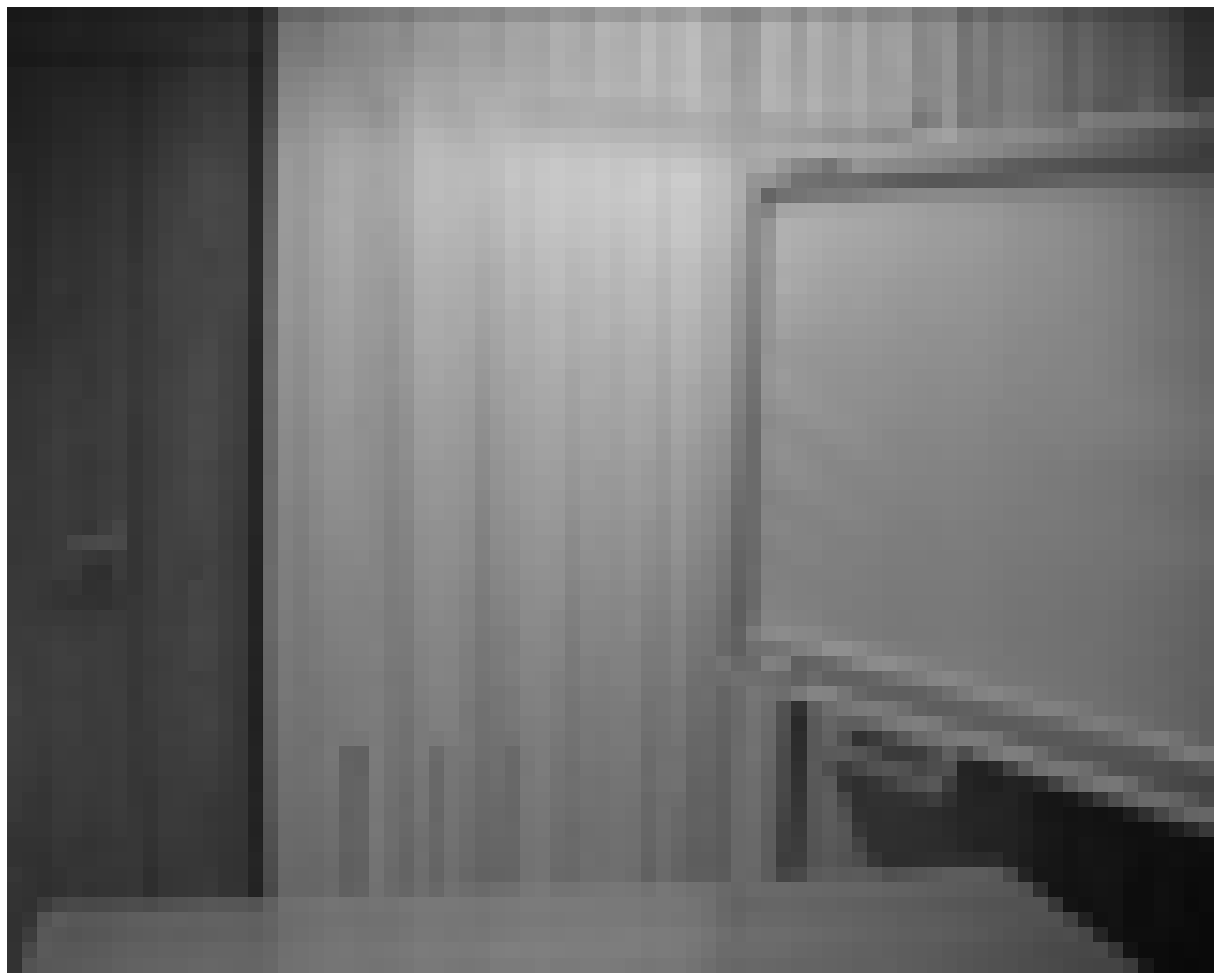}
		\includegraphics[width=25mm]{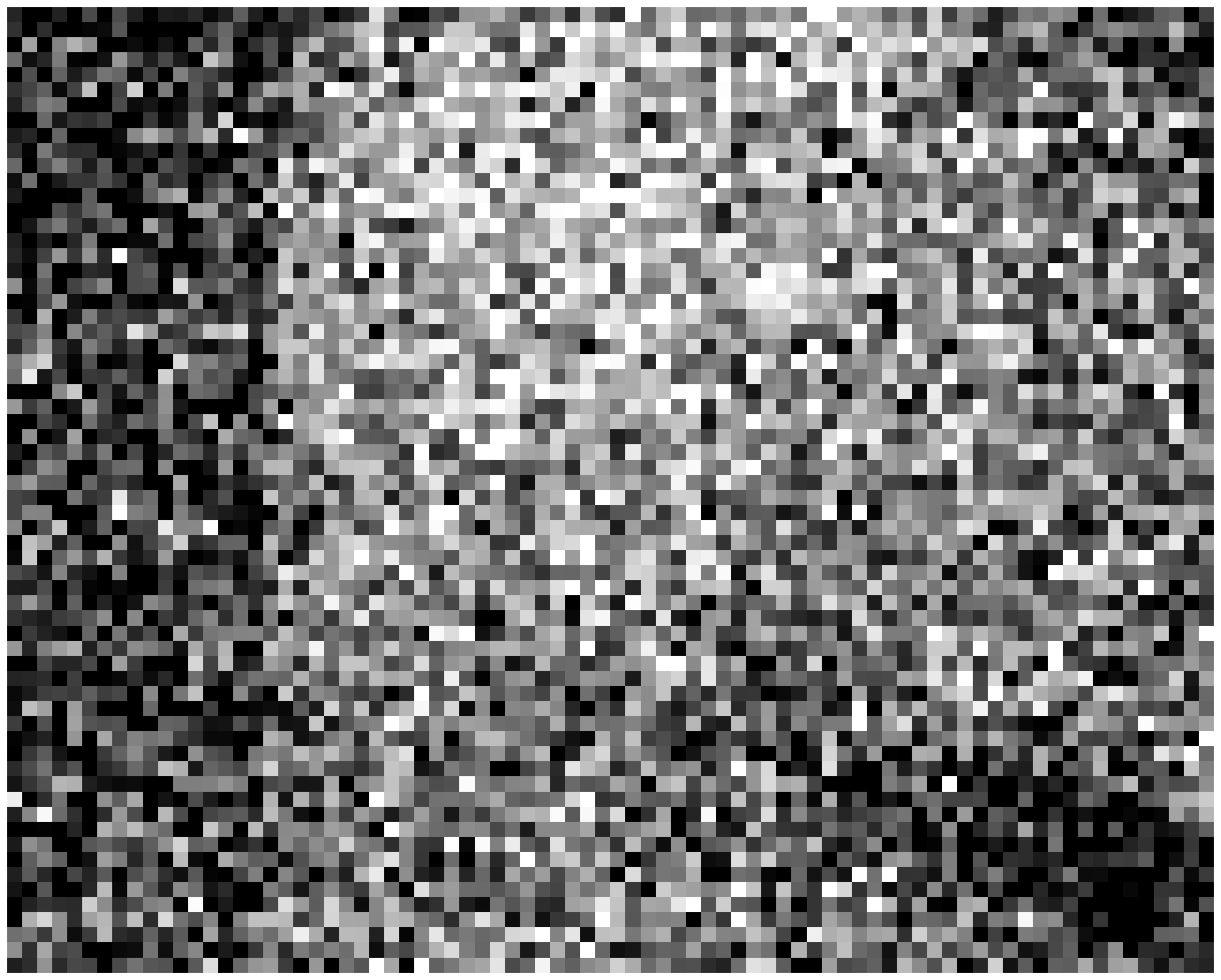}		
		\includegraphics[width=25mm]{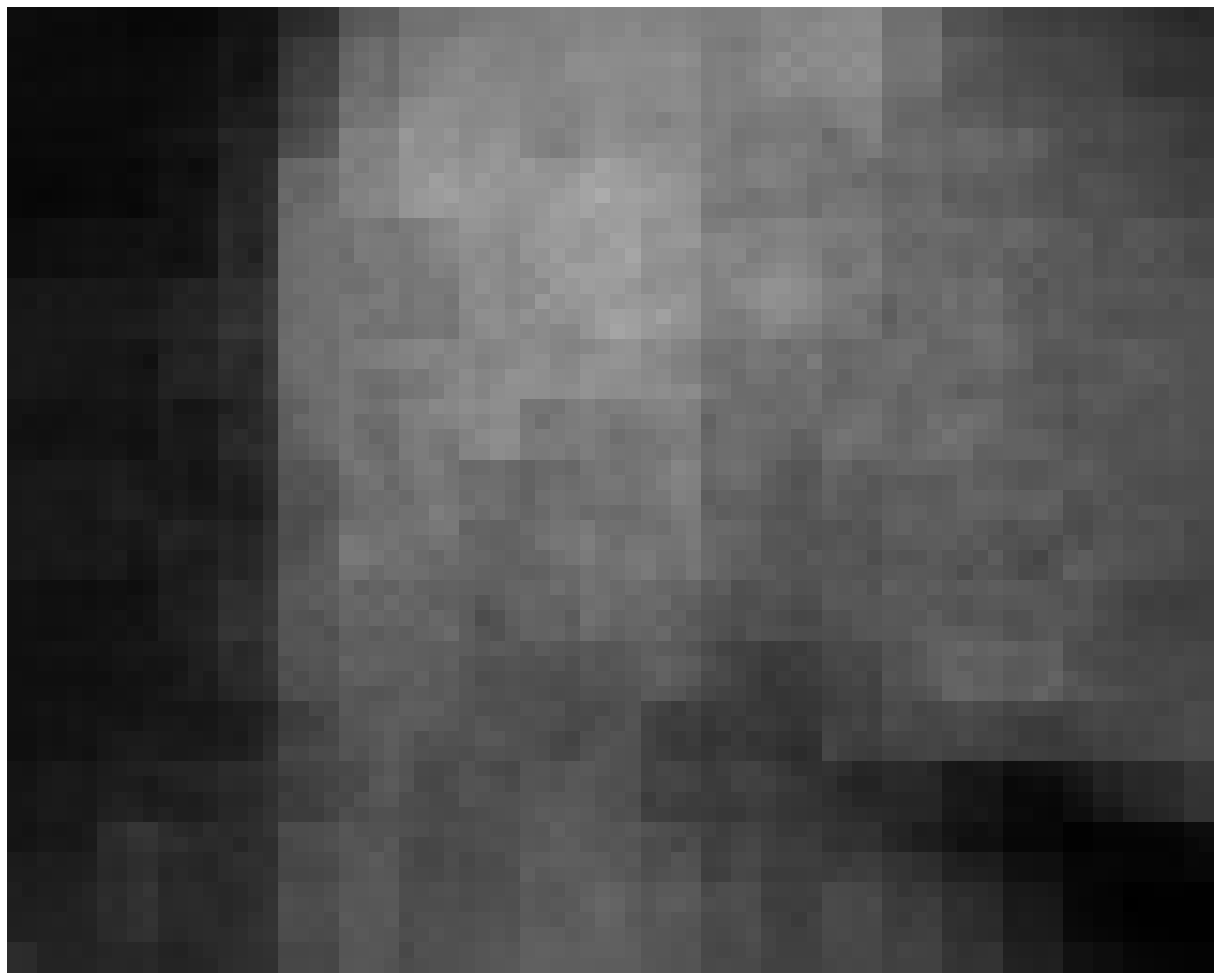}\\
		\hspace*{\fill}\makebox[0pt]{original }\hspace*{\fill}
		\hspace*{\fill}\makebox[0pt]{noisy}\hspace*{\fill}
		\hspace*{\fill}\makebox[0pt]{SLMA}\hspace*{\fill}\\
		\includegraphics[width=25mm]{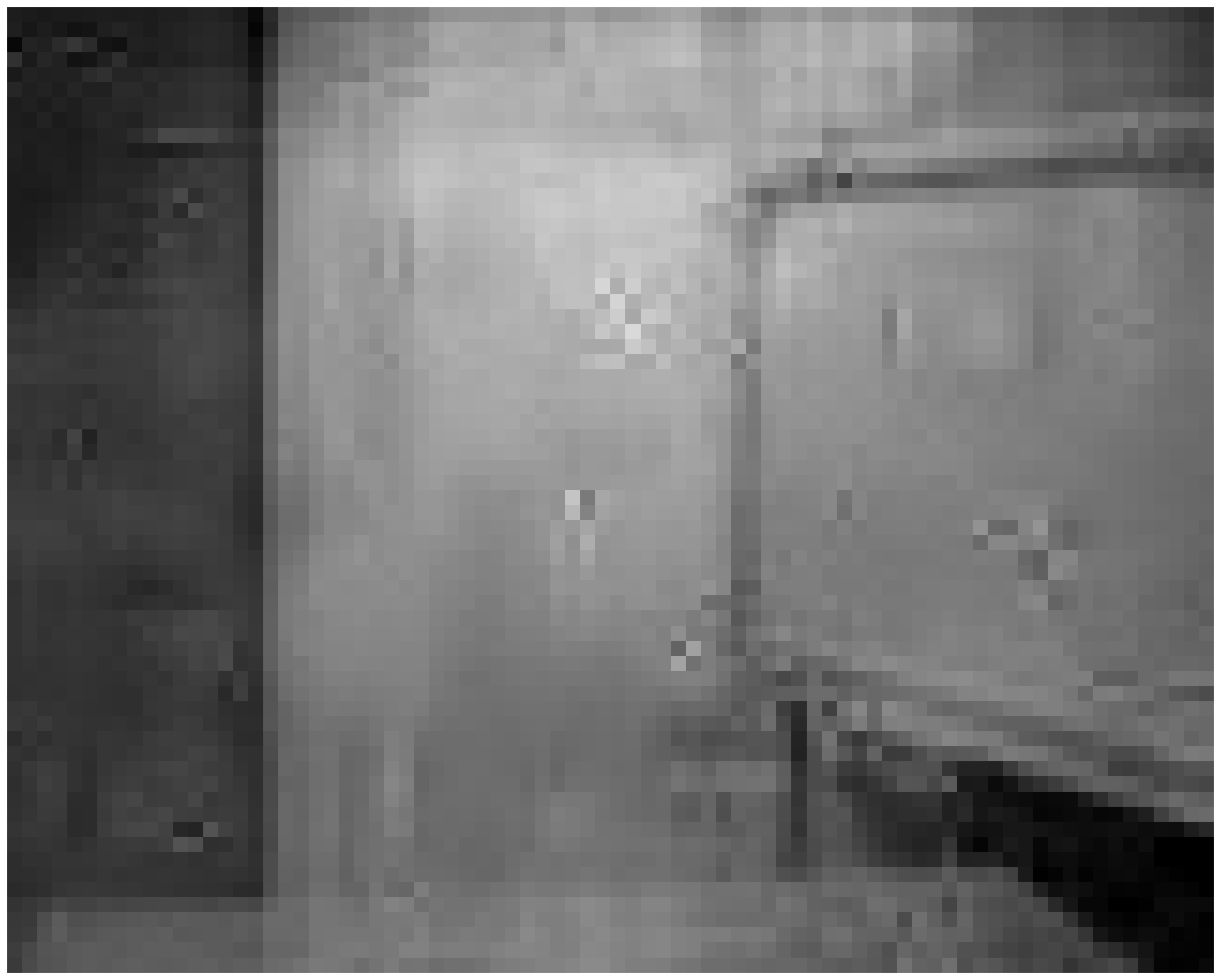}
		\includegraphics[width=25mm]{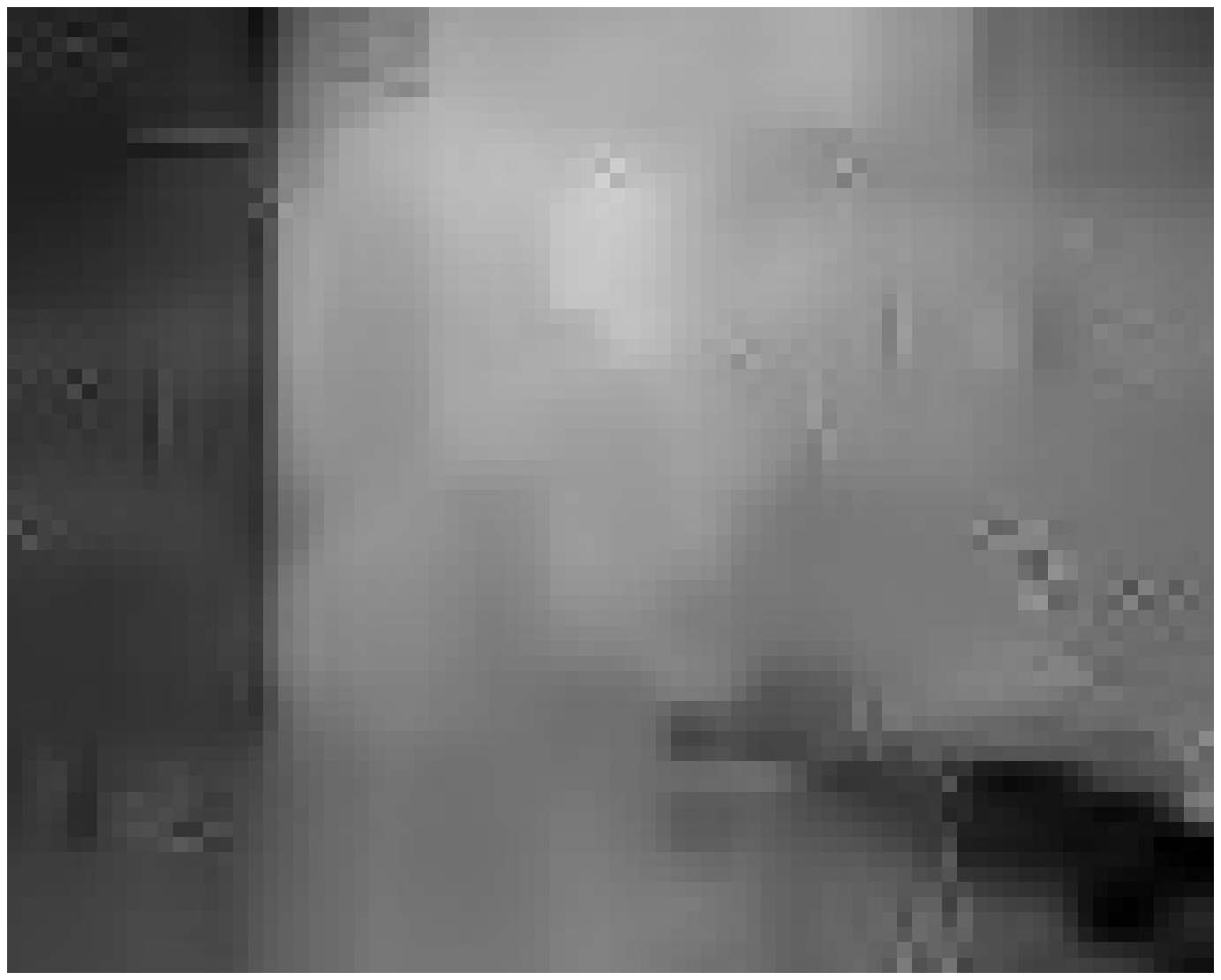}
		\includegraphics[width=25mm]{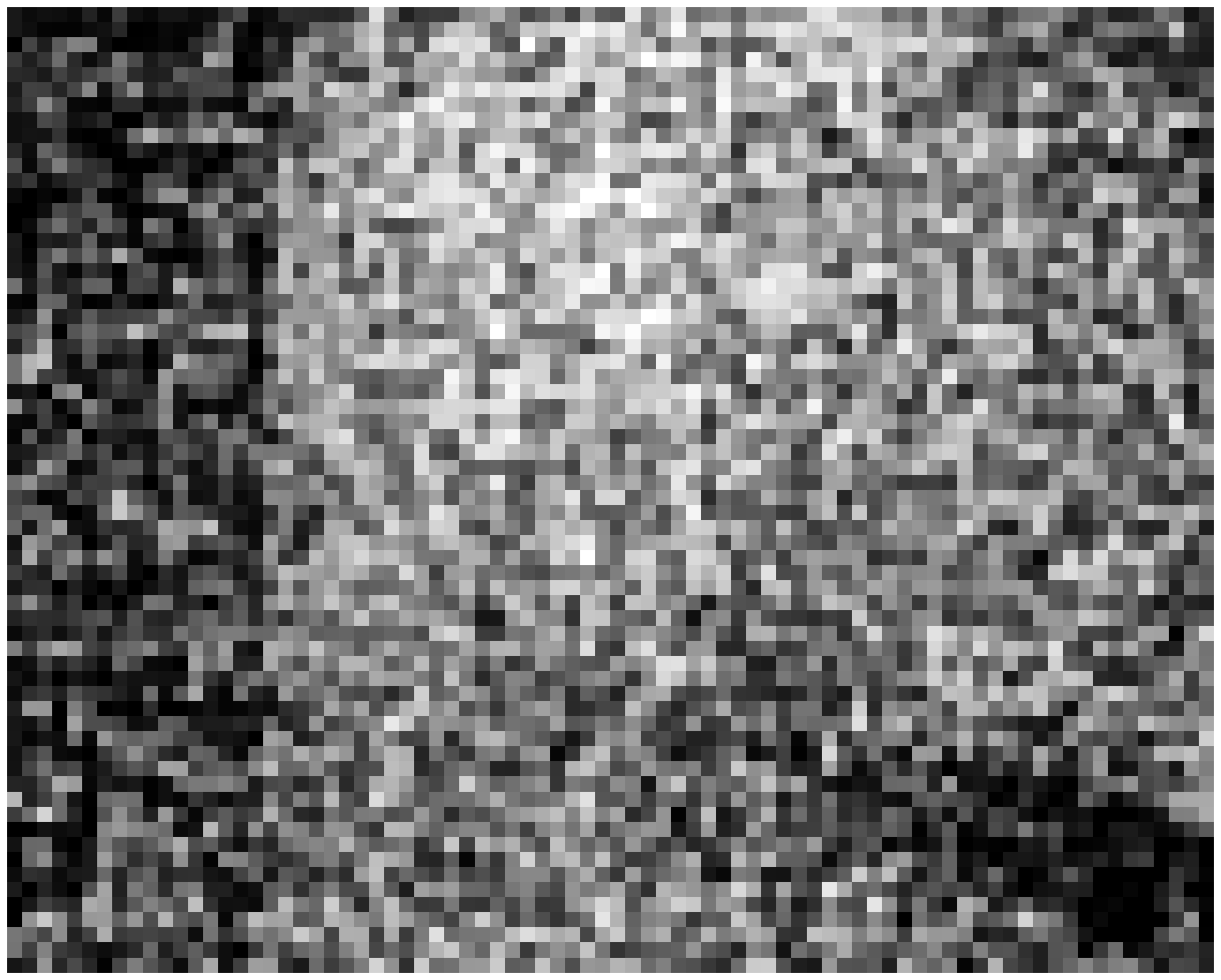}\\
		\hspace*{\fill}\makebox[0pt]{ReLD}\hspace*{\fill}
        \hspace*{\fill}\makebox[0pt]{VBM3D}\hspace*{\fill}
        \hspace*{\fill}\makebox[0pt]{MLP}\hspace*{\fill}

	\end{tabular}
	\caption{\small{Visual comparison of denoising performance for the Curtain dataset for very large Gaussian noise ($\sigma=70$)}\label{VisualCurtain}}
	\vspace{-0.3cm}	
\end{figure}

\subsection{Denoising in Low-light Environment}

\begin{table}
\centering
\begin{tabular}{| c | c | c | c | c | }
\hline
\multirow{2}{*}{$\sigma$}  & \multicolumn{4}{ c |}{Dataset: waterfall}  \\ \cline{2-5}
& ReLD & VBM3D & MLP & PCP-LD  \\
\hline
50 & 33.08(73.14) & 27.99(24.14) & 18.87(477.60) & 32.93(195.77) \\
\hline
70 & 29.25(69.77) & 24.42(21.01) & 15.03(478.73) & 29.17(197.94)\\
\hline
$\sigma$ &  \multicolumn{2}{c |}{NCRPCA-LD} & \multicolumn{2}{c|}{GRASTA-LD}   \\
\hline
50 & \multicolumn{2}{c|}{30.48(128.35)} & \multicolumn{2}{c|}{25.33(58.23)}  \\
\hline
70 & \multicolumn{2}{c|}{27.97(133.53)} & \multicolumn{2}{c|}{21.89(55.45)} \\
\hline
\hline
\multirow{2}{*}{$\sigma$}  & \multicolumn{4}{ c |}{Dataset: fountain}  \\ \cline{2-5}
& ReLD & VBM3D & MLP & SLMA  \\
\hline
50 & 30.53(15.82) & 26.55(5.24) & 18.53(109.79) & 18.55($3.13\times 10^4$)\\
\hline
70 & 27.53(15.03) & 22.08(4.69) & 14.85(107.52) & 16.25($3.19\times 10^4$)\\
\hline
\hline

\multirow{2}{*}{$\sigma$}  & \multicolumn{4}{ c |}{Dataset: escalator}  \\ \cline{2-5}
& ReLD & VBM3D & MLP & SLMA  \\
\hline
50 & 27.84(16.03) & 25.10(5.27) & 18.83 (109.40)& 17.98($3.21\times 10^4$) \\
\hline
70 & 25.15(15.28) & 20.20(4.72)& 15.20(108.78) & 15.90($3.18\times 10^4$)\\
\hline
\hline

\multirow{2}{*}{$\sigma$}  & \multicolumn{4}{ c |}{Dataset: curtain}  \\ \cline{2-5}
& ReLD & VBM3D & MLP & SLMA  \\
\hline
50 & 31.91(17.17) & 30.29(4.42) & 18.58(188.30) & 19.12($7.86\times 10^4$)\\
\hline
70 & 28.10(16.50) & 26.15(3.85) & 14.73(192.00) & 16.68($8.30\times 10^4$) \\
\hline
\hline

\multirow{2}{*}{$\sigma$}  & \multicolumn{4}{ c |}{Dataset: lobby}  \\ \cline{2-5}
& ReLD & VBM3D & MLP & SLMA  \\
\hline
50 & 35.15(58.41) & 29.23(19.35) & 18.66(403.59 )& 18.21($3.99\times 10^5$) \\
\hline
70 & 29.68(56.51) & 24.90(17.00) & 14.85(401.29) & 16.82($4.09\times 10^5$)\\
\hline

\end{tabular}
\caption{\small{PSNR (running time in seconds) for different denoising algorithms. The running time for MLP does not include training time.}}\label{TabPSNR}
\end{table}

\begin{figure}[!h]
	\centering
	\def\hei{10}
	\def\spa{2mm}
	\begin{tabular}{cc}
		\put(0,\hei){\makebox[0pt][r]{img$_1$\rule{\spa}{0pt}}}
		\includegraphics[width=23mm]{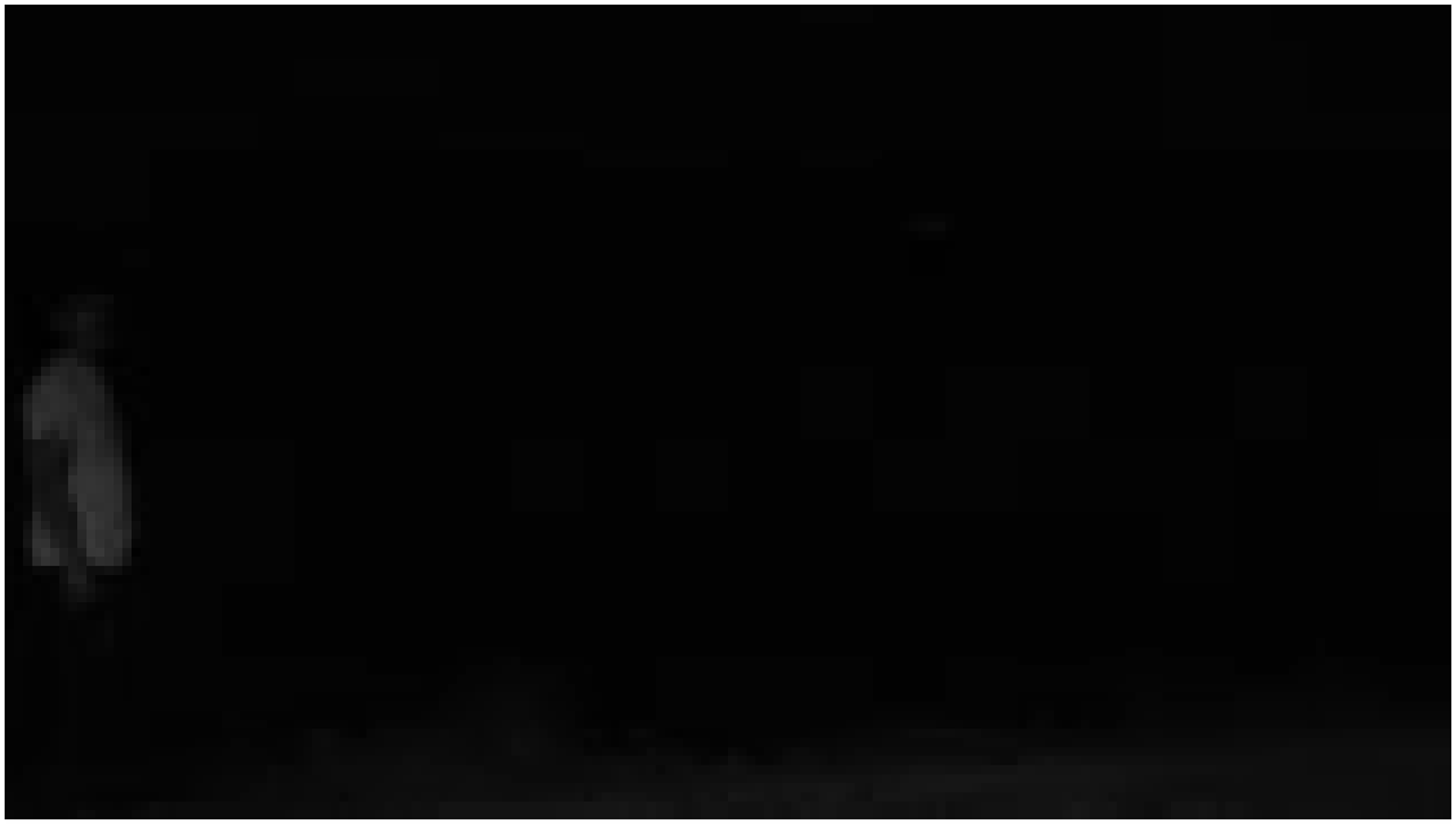}
		\includegraphics[width=23mm]{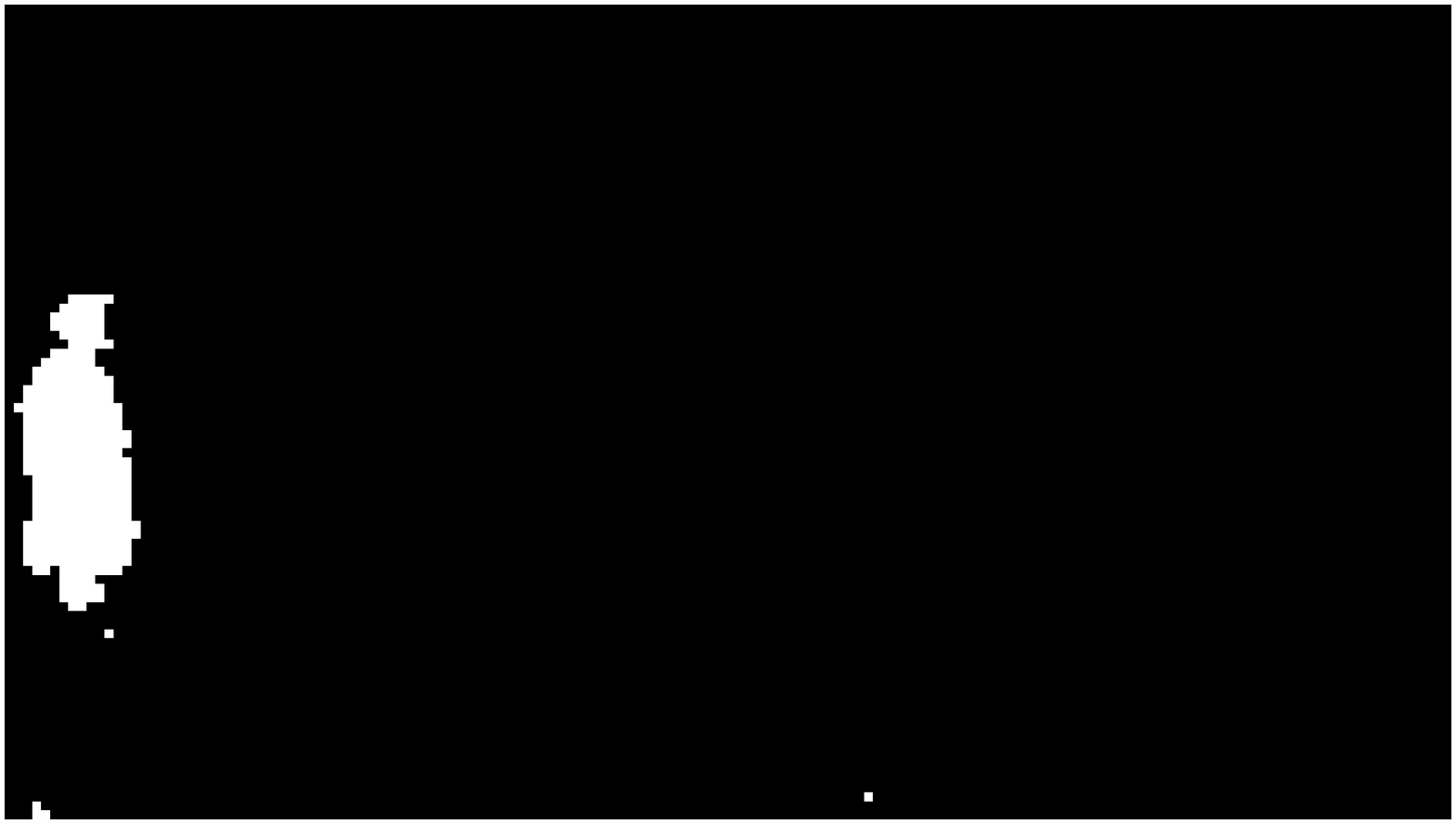}
		\includegraphics[width=23mm]{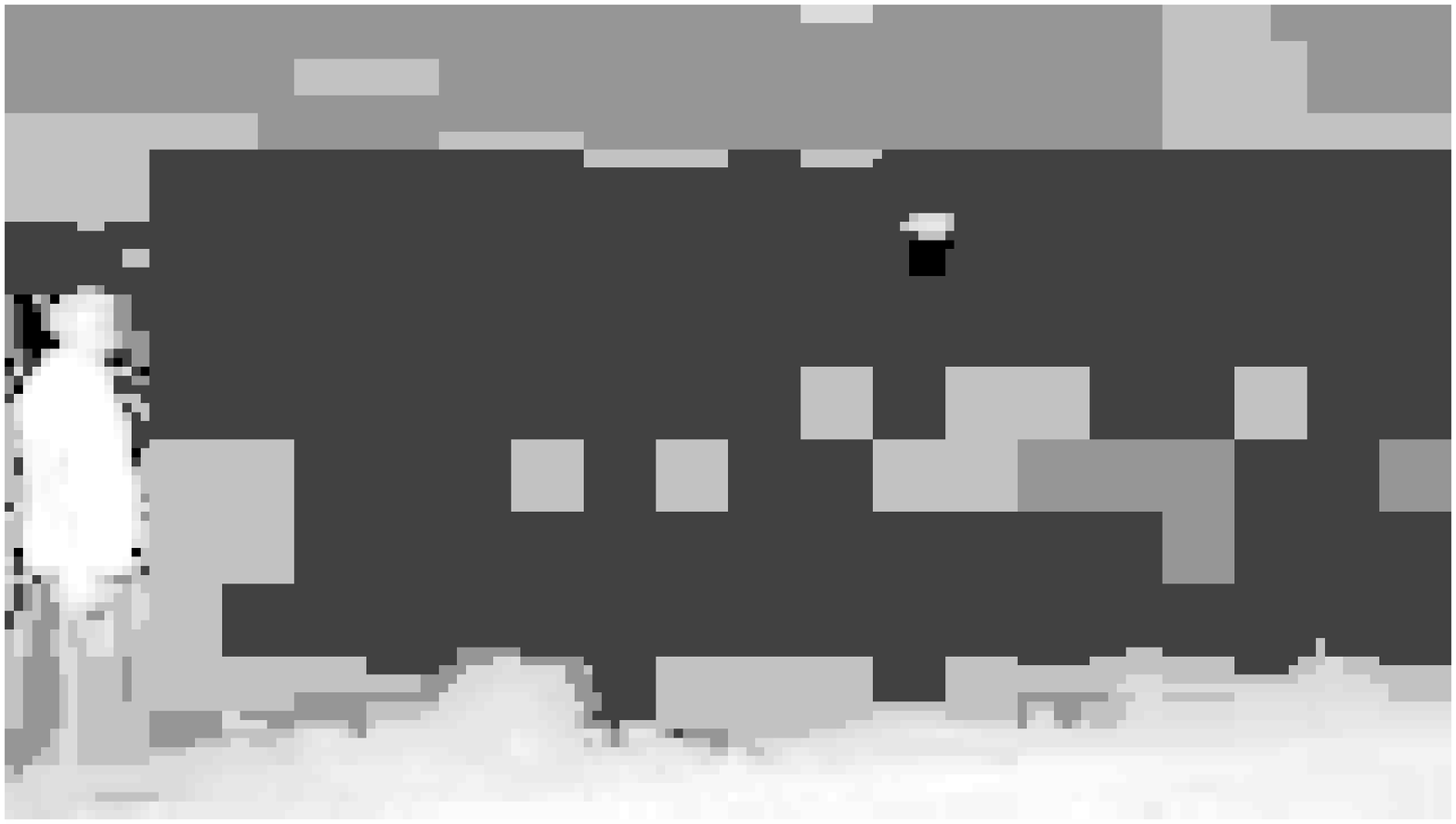}\\
		\put(0,\hei){\makebox[0pt][r]{img$_2$\rule{\spa}{0pt}}}
		\includegraphics[width=23mm]{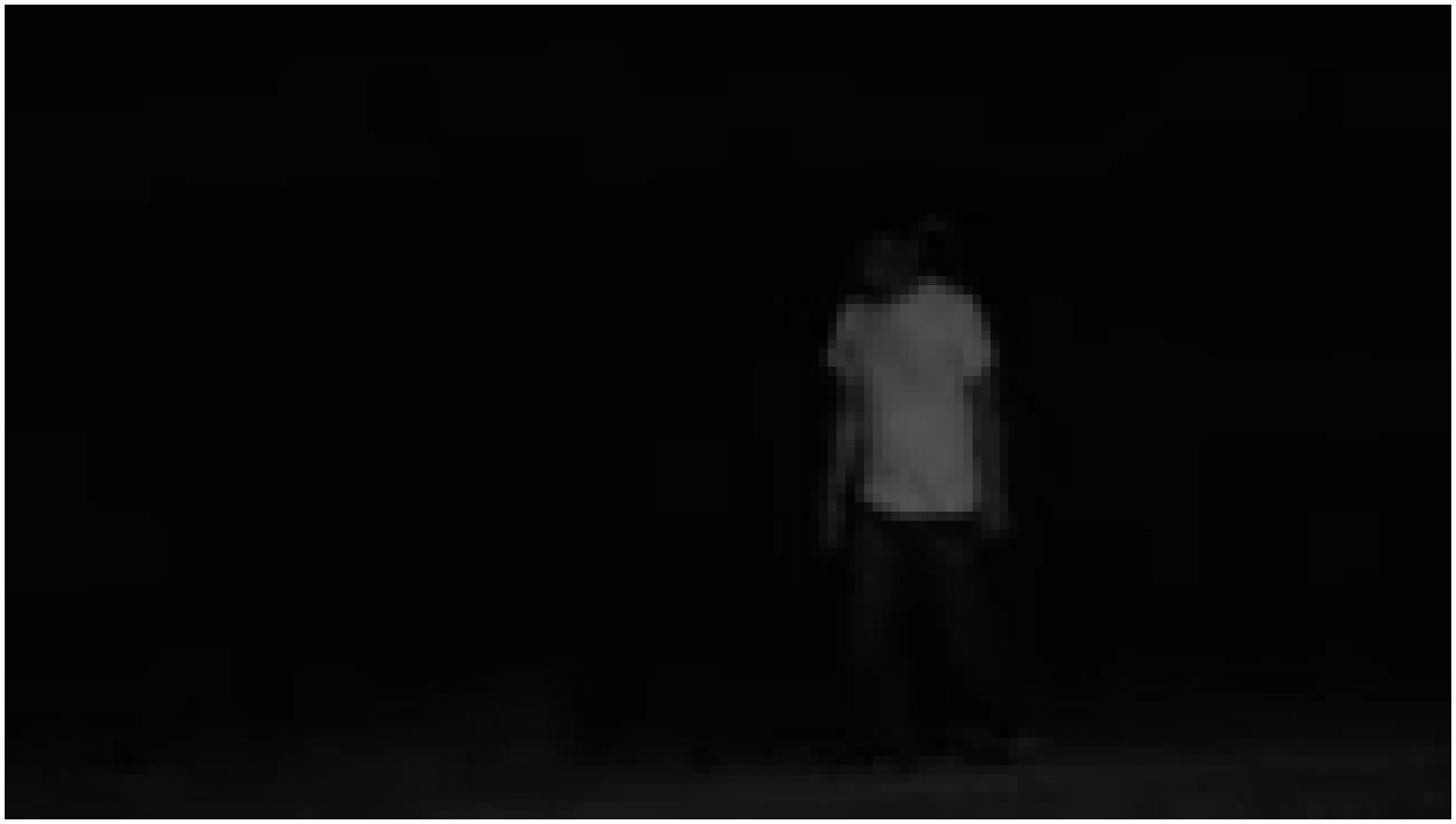}
		\includegraphics[width=23mm]{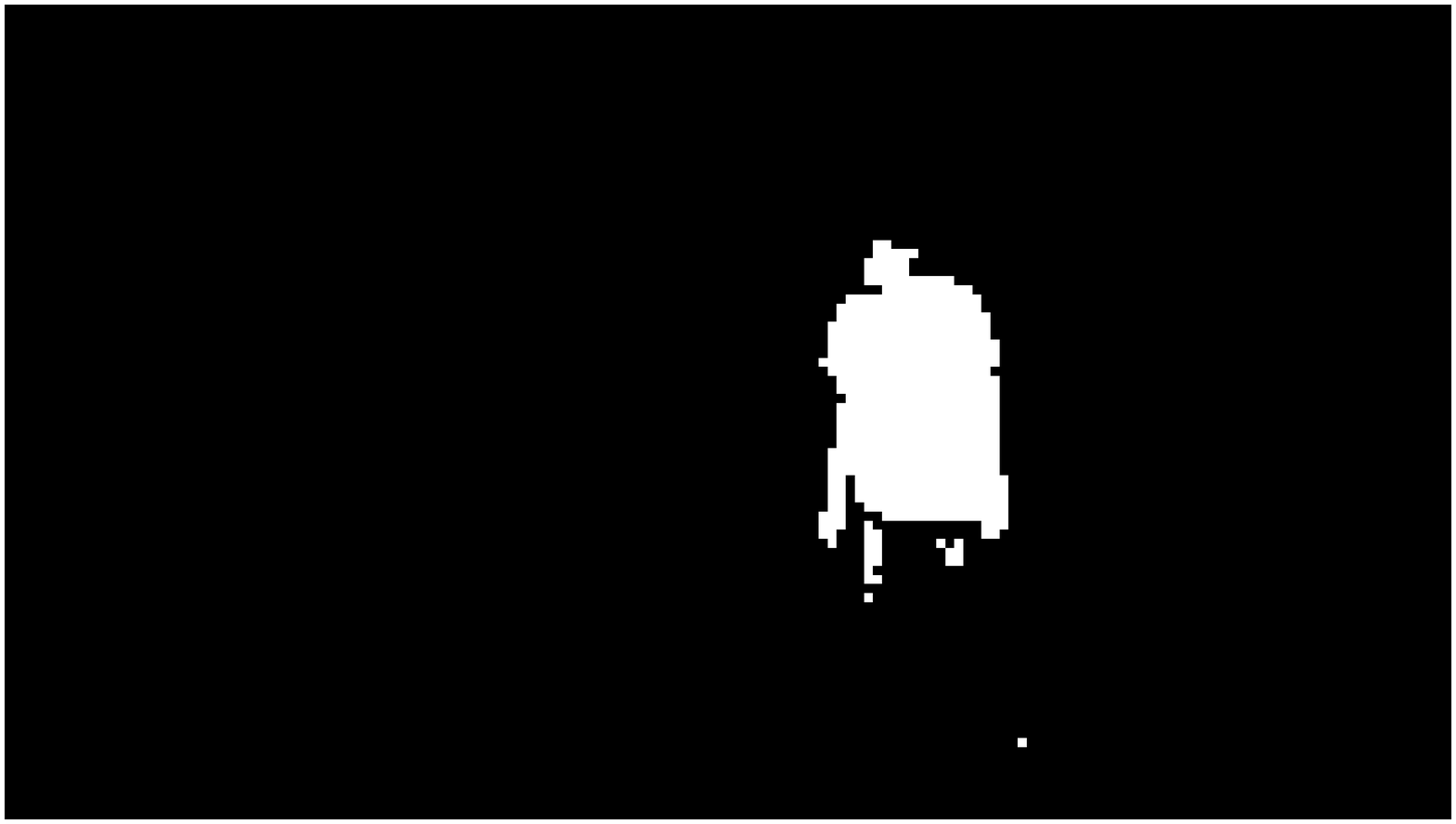}
		\includegraphics[width=23mm]{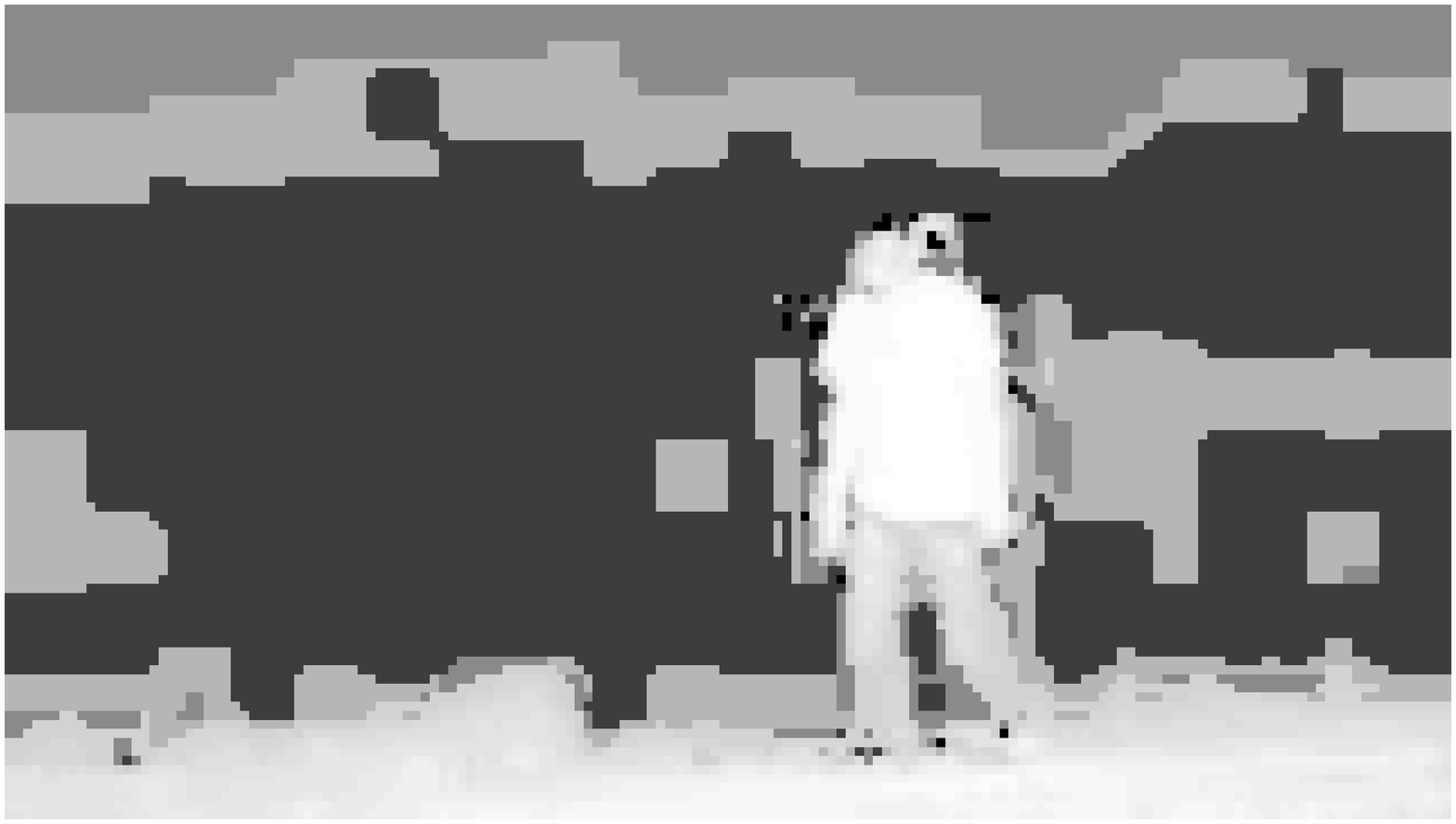}\\

	    \hspace*{\fill}\makebox[0pt]{original }\hspace*{\fill}
		\hspace*{\fill}\makebox[0pt]{ReLD}\hspace*{\fill}
		\hspace*{\fill}\makebox[0pt]{Hist-Eq}\hspace*{\fill}
%
	\end{tabular}\caption{\small{Ability of ``seeing'' in the dark for two sample frames. From left to right: orignal dark image, results using ReLD, and using Histogram-Equalization (Hist-Eq).}}\label{dark}
\end{figure}

In this part we show how ReLD can also be used for video enhancement of low-light videos, i.e., to see the target signal in the low-light environment. The video was taken in a dark environment where a barely visible person walked through the scene. The output we are using here for ReLD is $\hat{\bm{S}}$, which is the output of ReProCS. In Fig. \ref{dark} we see ReLD is able to enhance the visual quality -- observing the walking person. Observe that Histogram-Equalization, which is a standard technique for enhancing low-light images, does not work for this video.

 \section{Conclusion}

In this paper we developed a denoising scheme, called ReLD which enhances the denoising performance of the state-of-the-art algorithm VBM3D, and is able to achieve denoising in a broad sense. Using ReProCS to split the video first results in a clean low-rank layer since the large noise goes to the sparse layer. The clean layer improves the ``grouping'' accuracy in VBM3D. One drawback of our algorithm is that VBM3D needs to be executed twice, once on each layer. As a result, the running time is at least doubled, but that also results in significantly improved PSNRs especially for very large variance Gaussian noise. Moreover ReLD is still 6-10 times faster than SLMA and MLP, while also being better. It is also much faster than PCP-LD and NCRPCA-LD.

\vfill\pagebreak
\clearpage

\bibliographystyle{IEEEbib}

\bibliography{tipnewpfmt_kfcsfullpap}

\clearpage
\pagebreak
\onecolumn

\begin{center}
{\huge{\textbf{Supplementary Material}}}
\end{center}

\section*{Complete Algorithm}

Detailed algorithm of ReLD is summarized in Algorithm \ref{Alg2}.

\begin{algorithm*}
\caption{\small{Details of each step of ReLD}}\label{Alg2}
\textbf{Parameters:} We used $\alpha = 20, K_{\min}=3, K_{\max}=10, t_0=50$ in all experiments.
\begin{enumerate}
\item Initialization using PCP: Compute $(\hat{\bm{L}}_0, \hat{\bm{S}}_0) \leftarrow \text{PCP}(\bm{M}_0)$ and compute $[\hat{\bm{P}}_0,\hat{\bm{\Sigma}}_0] \leftarrow \text{approx-basis}(\hat{\bm{L}}_0,90\%)$. The notation PCP$(\bm{M})$ means implementing the PCP algorithm on matrix $\bm{M}$ and $\bm{P}$ = \text{approx-basis}$(\bm{M},b\%)$ means that $\bm{P}$ is the $b\%$ left singular vectors' matrix for $\bm{M}$.\\
Set $\hat{r}\leftarrow \text{rank}(\bm{\hat{P}}_0)$, $\hat{\bm{\sigma}}_{\min}\leftarrow (\bm{\hat{\Sigma}}_0)_{\hat{r},\hat{r}}$, $\hat{t}_0 =t_0$, flag$=$detect

\item For all $t > t_0$, implement an appropriately modified ReProCS algorithm
\begin{enumerate}
\item  Split $\bm{m}_t$ into layers $\bm{\hat{\ell}}_t$ and $\bm{\hat{s}}_t$:
\begin{enumerate}
\item Compute $\bm{y}_t\leftarrow  \bm{\Phi}_t \bm{m}_t$ with $\bm{\Phi}_t \leftarrow \bm{I} - \hat{\bm{P}}_{t-1}\hat{\bm{P}}_{t-1}'$
\item Compute $\hat{\bm{s}}_t$ as the solution of $$ \min_{\bm{x}} \| \bm{x} \|_1 \text{s.t.} \|   \bm{y}_t -  \bm{\Phi}_t  \bm{x}  \|_2  \leq \xi $$
with $\xi =\| \bm{\Phi}_t \hat{\bm{\ell}}_{t-1} \|$
\item $\hat{\T}_t \leftarrow \text{Thresh}(\bm{\hat{s}_t}, \omega)$ with $\omega = 3\sqrt{\| \bm{m}_t \|^2/n}$. Here $\T\leftarrow \text{Thresh}(\bm{x},\omega)$ means that $\T = \left\lbrace i: |(\bm{x})_i |\geq \omega \right\rbrace$\\
$\hat{\bm{s}}_{t,\ast} \leftarrow \text{LS}(\bm{y}_t, \bm{\Phi}_t, \hat{\T}_t)$. Here $\hat{\bm{x}}\leftarrow \text{LS}(\bm{y},\bm{A},\T)$ means that $\hat{\bm{x}}_\T = (\bm{A}_\T'\bm{A}_\T)^{-1}\bm{A}_\T'\bm{y}$, which is least-squared estimate of $\bm{x}$ on $\T$.
\item $\hat{\bm{\ell}}_t \leftarrow \bm{m}_t - \hat{\bm{s}}_t$, $\hat{\bm{\ell}}_{t,\ast} \leftarrow \bm{m}_t - \hat{s}_{t,\ast}$
\end{enumerate}

\item Perform subspace update, i.e., update $\hat{\bm{P}}_t$:

\ben
\item If $\text{flag} = \text{detect}$ and  $\text{mod}(t - \hat{t}_j +1, \alpha)=0$,

\ben
\item  compute the SVD of $ \frac{1}{\sqrt{\alpha}}(\bm{I} - \hat{\bm{P}}_{(j-1)} {\hat{\bm{P}}_{(j-1)}}')[{\hat{\bm{\ell}}}_{t-\alpha+1,\ast}, \dots {\hat{\bm{\ell}}}_{t,\ast}]$ and check if any singular values are above $\hat\sigma_{\min}$
\item if the above number is more than zero then set $\text{flag} \leftarrow \text{pPCA}$, increment $j \leftarrow j+1$, set $\hat{t}_j \leftarrow t-\alpha+1$, reset $k \leftarrow 1$
\een

Else $\hat{\bm{P}}_{t}  \leftarrow \hat{\bm{P}}_{t-1}$.

\item If $\text{flag} = \text{pPCA}$ and  $\text{mod}(t - \hat{t}_j +1, \alpha)=0$,

\ben
\item compute the SVD of $ \frac{1}{\sqrt{\alpha}}(\bm{I} - \hat{\bm{P}}_{j-1} {\hat{\bm{P}}_{(j-1)}}')[\hat{\bm{\ell}}_{t-\alpha+1,\ast}, \dots \hat{\bm{\ell}}_{t,\ast}]$,
\item  let $\hat{\bm{P}}_{j,\new,k}$ retain all its left singular vectors with singular values above $\hat\sigma_{\min}$ or all $\alpha/3$ top left singular vectors whichever is smaller,
\item update $\hat{\bm{P}}_{t} \leftarrow [\hat{\bm{P}}_{(j-1)} \ \hat{\bm{P}}_{j,\new,k}]$, increment $k \leftarrow k+1$

\item If $k \geq  K_{\min}$ and $\frac{\|\sum_{t-\alpha+1}^t (\hat{\bm{P}}_{j,\new,i-1} {\hat{\bm{P}}_{j,\new,i-1}}' - \hat{\bm{P}}_{j,\new,i} {\hat{\bm{P}}_{j,\new,i}}')\hat{\bm{\ell}}_{t,\ast}\|_2}{\|\sum_{t-\alpha+1}^t \hat{\bm{P}}_{j,\new,i-1} {\hat{\bm{P}}_{j,\new,i-1}}' \hat{\bm{\ell}}_{t,\ast}\|_2 }< 0.01$ for $i=k-2,k-1,k$; or $k = K_{\max}$,
\\    then $K \leftarrow  k$, $\hat{\bm{P}}_{(j)} \leftarrow [\hat{\bm{P}}_{(j-1)} \ \hat{\bm{P}}_{j,\new,K}]$ and reset $\text{flag} \leftarrow \text{detect}$. 
\label{stop_crit}
\een

Else $\hat{\bm{P}}_t  \leftarrow \hat{\bm{P}}_{t-1}$.

\een

\end{enumerate}

\item Denoise using VBM3D:

\begin{enumerate}
\item $\hat{\sigma}_{\text{fg}} \leftarrow \text{Std-est}([\hat{\bm{s}}_{t},\ldots, \hat{\bm{s}}_{t_0}])$\\
$\hat{\sigma}_{\text{bg}} \leftarrow \text{Std-est}([\hat{\bm{\ell}}_{t},\ldots, \hat{\bm{\ell}}_{t_0}])$. Here Std-est$(\bm{M})$ denotes estimating the standard deviation of noise from $\bm{M}$: we first subtract column-wise mean from $\bm{M}$ and then compute the standard deviation by seeing it as a vector.

\item $\hat{\bm{S}}_{\text{denoised}}\leftarrow \text{VBM3D}([\hat{\bm{s}}_{1},\ldots, \hat{\bm{s}}_{t_{\text{max}}}], \hat{\sigma}_{\text{fg}} )$\\
$\hat{\bm{L}}_{\text{denoised}}\leftarrow \text{VBM3D}([\hat{\bm{\ell}}_{1},\ldots, \hat{\bm{\ell}}_{t_{\text{max}}}],\hat{\sigma}_{\text{bg}})$. Here \text{VBM3D}$(\bm{M}, \sigma)$ implements the VBM3D algorithm on matrix $\bm{M}$ with input standard deviation~$\sigma$.

\end{enumerate}

\end{enumerate}
\textbf{Output:} $\hat{\bm{S}}$, $\hat{\bm{S}}_{\text{denoised}}$, $\hat{\bm{L}}_{\text{denoised}}$ or $\hat{\bm{I}}_{\text{denoised}}=\hat{\bm{S}}_{\text{denoised}}+\hat{\bm{L}}_{\text{denoised}}$ based on applications
\end{algorithm*}

\section*{Complete Experiments}

\begin{table*}
\centering
\begin{tabular}{| c  | c | c | c | c| c| c| }
\hline

$\sigma$ & ReLD & PCP-LD & NCRPCA-LD & GRASTA-LD & VBM3D & MLP\\
\hline

25 & 35.00, 32.78 (73.54)  & 34.92, 32.84 (198.87) & 33.34, 31.98 (101.78) & 30.45, 28.11 (59.43) & 32.02 (24.83) & 28.26 (477.22)  \\
\hline

30 & 34.51, 32.68 (73.33)  & 34.42, 32.60 (185.47) & 32.53, 31.56 (106.30) & 29.40, 26.89 (58.76) & 30.96 (23.96) & 26.96 (474.26)  \\
\hline

50 & 33.08, 32.27 (73.14)  & 32.93, 31.65 (195.77) & 30.48, 30.09 (128.35) & 25.33, 23.97 (58.23) & 27.99 (24.14) & 18.87 (477.60)  \\
\hline

70 & 29.25, 31.79 (69.77) &  29.17, 30.67 (197.94) & 27.97, 29.63 (133.53) & 21.89, 21.81 (55.45) & 24.42 (21.01) & 15.03 (478.73)  \\
\hline

\end{tabular}

\caption{Comparison of denoising performance (PSNR) on the Waterfall dataset (length 650 and the images are of size $108 \times 192$): data shown in format of PSNR using $\hat{\bm{I}}_{\text{denoised}}$, PSNR using $\hat{\bm{L}}_{\text{denoised}}$,  (and runing time in second).}\label{WaterfallSmall}
\end{table*}

\begin{table*}
\centering
\begin{tabular}{| c | c | c | c | c| c| c|c| }
\hline

$\sigma$ & ReLD  & PCP-LD & NCRPCA-LD & GRASTA-LD & VBM3D & MLP\\
\hline
25 & 33.84, 29.98 (335.97)  & 33.38, 29.17 (413.45) & 28.99, 27.54 (507.03) & 29.37, 16.12 (435.54) & 33.67 (110.89) & 31.11 ($1.38\times 10^3$)\\
\hline
30 & 33.01, 29.79 (338.51)  & 32.49, 28.72 (406.45) & 27.63, 27.63 (510.39) & 28.16, 12.53 (405.53) & 32.75 (110.92) & 29.18 ($1.38\times 10^3$)\\
\hline
50 & 30.48, 28.86 (333.26)  & 29.79, 26.83 (399.09) & 23.74, 23.29 (544.60) & 22.17, 11.27 (417.50) & 30.18 (110.83) & 19.00 ($1.39\times 10^3$)\\
\hline
70 & 27.39, 27.77 (321.71)  & 26.80, 25.06 (386.60) & 20.95, 20.88 (589.22) & 13.92, 9.78 (411.87) & 26.75 (105.80) & 15.08 ($1.40\times 10^3$)\\
\hline
\end{tabular}

\caption{Comparison of denoising performance (PSNR) on the Waterfall dataset (length 100 and the images are of size $540 \times 960$): data shown in format of PSNR using $\hat{\bm{I}}_{\text{denoised}}$, PSNR using $\hat{\bm{L}}_{\text{denoised}}$,  (and runing time in second).}\label{WaterfallMid}

\end{table*}

\begin{table*}
\centering
\begin{tabular}{| c | c | c | c | c| c| c|c| }
\hline

$\sigma$ & ReLD & PCP-LD & NCRPCA-LD & GRASTA-LD & VBM3D & MLP\\
\hline
25 & 35.13, 29.79 ($2.83\times 10^3$)  & 34.52, 29.03 ($6.31\times 10^3$) & 30.69, 28.50 ($2.88\times 10^3$) & 29.94, 10.72 ($1.83\times 10^3$) & 36.04 (533) & 33.73 ($5.56\times 10^3$)\\
\hline
30 & 34.14, 29.61 ($2.34\times 10^3$)  & 33.45, 28.59 ($5.50\times 10^3$) & 29.22, 27.62 ($2.38\times 10^3$) & 28.46, 10.42 ($1.88\times 10^3$) & 35.18 (550) & 30.94 ($5.49\times 10^3$)\\
\hline
50 & 31.10, 28.72 ($2.31\times 10^3$)  & 30.31, 26.75 ($4.89\times 10^3$) & 25.18, 24.58 ($2.45\times 10^3$) & 22.27, 8.69 ($1.98\times 10^3$) & 32.55 (536) & 19.06 ($5.50\times 10^3$)\\
\hline
70 & 27.71, 27.66 ($2.38\times 10^3$)  & 27.08, 25.00 ($4.82\times 10^3$) & 22.23, 22.09 ($2.48\times 10^3$) & 13.59, 8.33 ($1.98\times 10^3$) & 28.45 (608) & 15.09 ($5.55\times 10^3$)\\
\hline
\end{tabular}

\caption{Comparison of denoising performance (PSNR) on the Waterfall dataset (length 100 and the images are of size $1080 \times 1920$): data shown in format of PSNR using $\hat{\bm{I}}_{\text{denoised}}$, PSNR using $\hat{\bm{L}}_{\text{denoised}}$,  (and runing time in second).}\label{WaterfallLarge}
\end{table*}

\begin{table*}
\centering
\begin{tabular}{| c | c | c | c | c | c | c | c | c |}
\hline
\multirow{2}{*}{$\sigma$}  & \multicolumn{4}{ c |}{Dataset: fountain} & \multicolumn{4}{ c |}{Dataset: escalator} \\ \cline{2-9}

& ReLD & VBM3D & MLP & SLMA & ReLD & VBM3D & MLP & SLMA \\
\hline

25 & 32.67(16.70) & 31.18(5.44) & 26.86(105.64) & 22.93($3.05\times 10^4$) & 31.01(16.64) & 30.32(5.34) & 25.53(107.51) & 21.17($3.09\times 10^4$) \\
\hline

30 & 32.25(15.84) & 30.26(5.17) & 25.67(107.41) & 21.85($3.06\times 10^4$) & 30.27(16.45) & 29.29(5.38) & 24.54(108.65) & 20.49($3.15\times 10^4$) \\
\hline

50 & 30.53(15.82) & 26.55(5.24) & 18.53(109.79) & 18.55($3.13\times 10^4$) & 27.84(16.03) & 25.10(5.27) & 18.83 (109.40)& 17.98($3.21\times 10^4$) \\
\hline

70 & 27.53(15.03) & 22.08(4.69) & 14.85(107.52) & 16.25($3.19\times 10^4$) & 25.15(15.28) & 20.20(4.72)& 15.20(108.78) & 15.90($3.18\times 10^4$)\\
\hline

\multirow{2}{*}{$\sigma$}  & \multicolumn{4}{ c |}{Dataset: curtain} & \multicolumn{4}{ c |}{Dataset: lobby} \\ \cline{2-9}

& ReLD & VBM3D & MLP & SLMA & ReLD & VBM3D & MLP & SLMA \\
\hline

25 & 35.47(16.78) & 34.60(4.15) & 31.14(189.14) & 23.28($7.75\times 10^4$) & 39.78(57.96) & 35.00(19.57) & 29.22(384.11) & 23.43($3.75\times 10^5$) \\
\hline

30 & 34.58(17.35) & 33.59(4.37) & 28.90(191.14) & 22.74($9.05\times 10^4$) & 38.76(57.99) & 33.64(19.09) & 27.72(395.67) & 21.15($3.82\times 10^5$) \\
\hline

50 & 31.91(17.17) & 30.29(4.42) & 18.58(188.30) & 19.12($7.86\times 10^4$) & 35.15(58.41) & 29.23(19.35) & 18.66(403.59 )& 18.21($3.99\times 10^5$) \\
\hline

70 & 28.10(16.50) & 26.15(3.85) & 14.73(192.00) & 16.68($8.30\times 10^4$) & 29.68(56.51) & 24.90(17.00) & 14.85(401.29) & 16.82($4.09\times 10^5$)\\
\hline
\end{tabular}
\caption{PSNR (and running time in second) for different denoising algorithms on datasets of fountain, escalator, curtain and lobby.}\label{TabPSNR}
\end{table*}

\begin{figure}
				\centering
				\begin{subfigure}{0.47\textwidth}
					\begin{tabular}{cc}
				\includegraphics[width=25mm]{Cur_Orig.eps}
				\includegraphics[width=25mm]{Cur_Noisy.eps}		
				\includegraphics[width=25mm]{Cur_SLMA.eps}\\
				\hspace*{\fill}\makebox[0pt]{original }\hspace*{\fill}
				\hspace*{\fill}\makebox[0pt]{noisy}\hspace*{\fill}
				\hspace*{\fill}\makebox[0pt]{SLMA}\hspace*{\fill}\\
				\includegraphics[width=25mm]{Cur_Alg1.eps}
				\includegraphics[width=25mm]{Cur_BM3D.eps}
				\includegraphics[width=25mm]{Cur_MLP.eps}\\
				\hspace*{\fill}\makebox[0pt]{ReLD}\hspace*{\fill}
        		\hspace*{\fill}\makebox[0pt]{VBM3D}\hspace*{\fill}
        		\hspace*{\fill}\makebox[0pt]{MLP}\hspace*{\fill}
					\end{tabular}
			
					\caption{Curtain}\label{VisualCurtain}
				\end{subfigure}
				\begin{subfigure}{0.47\textwidth}
					\begin{tabular}{cc}
					\includegraphics[width=25mm]{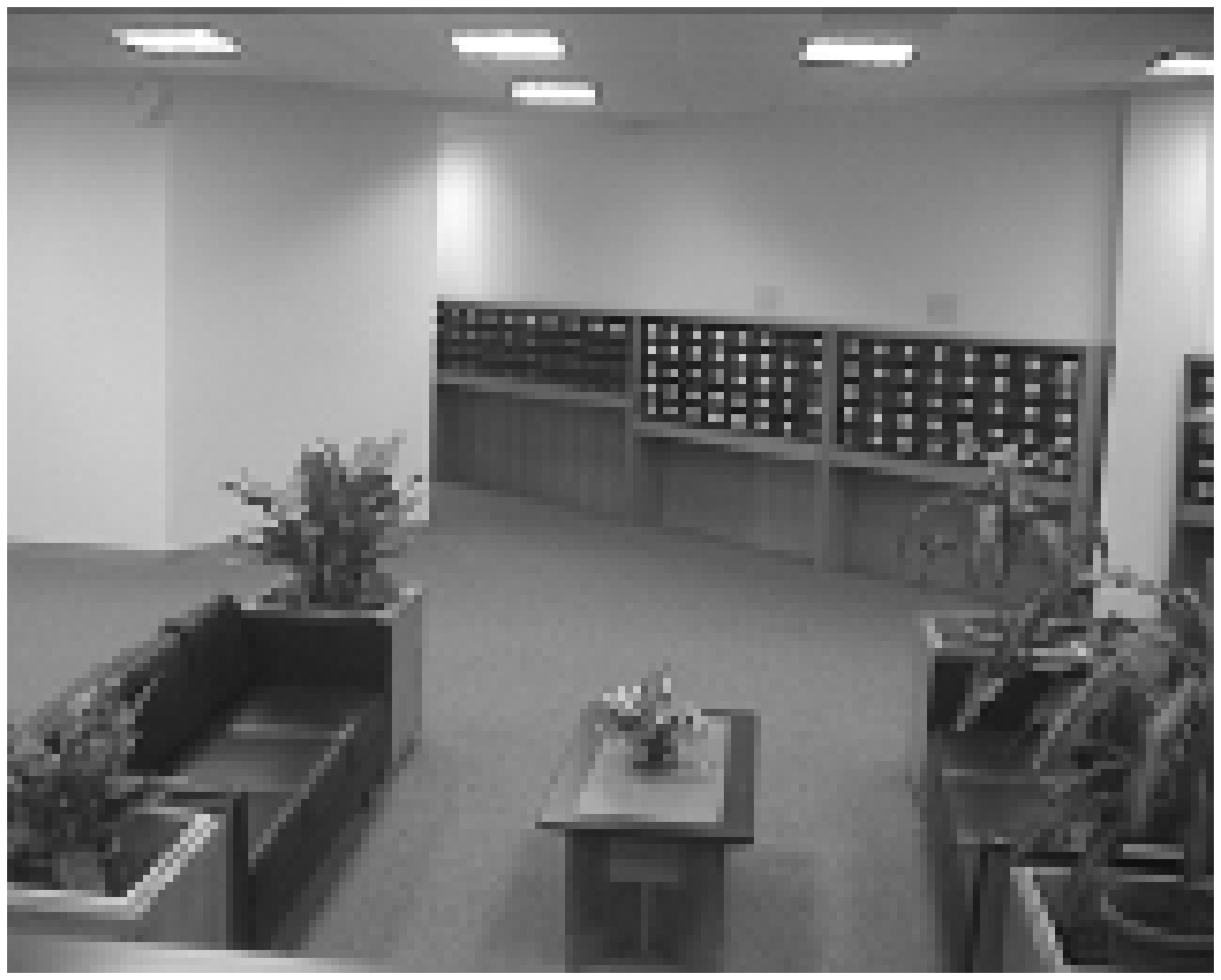}
					\includegraphics[width=25mm]{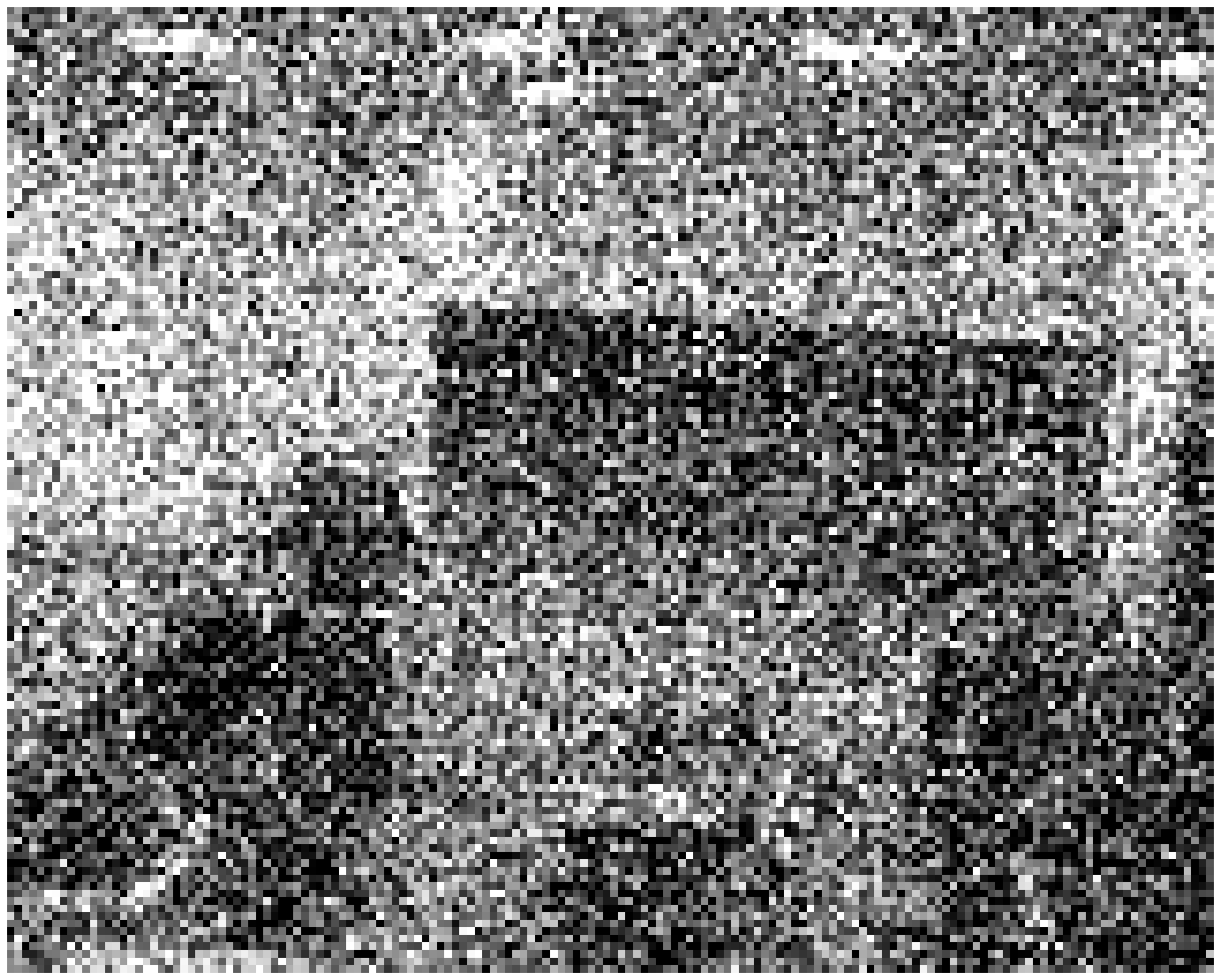}		
					\includegraphics[width=25mm]{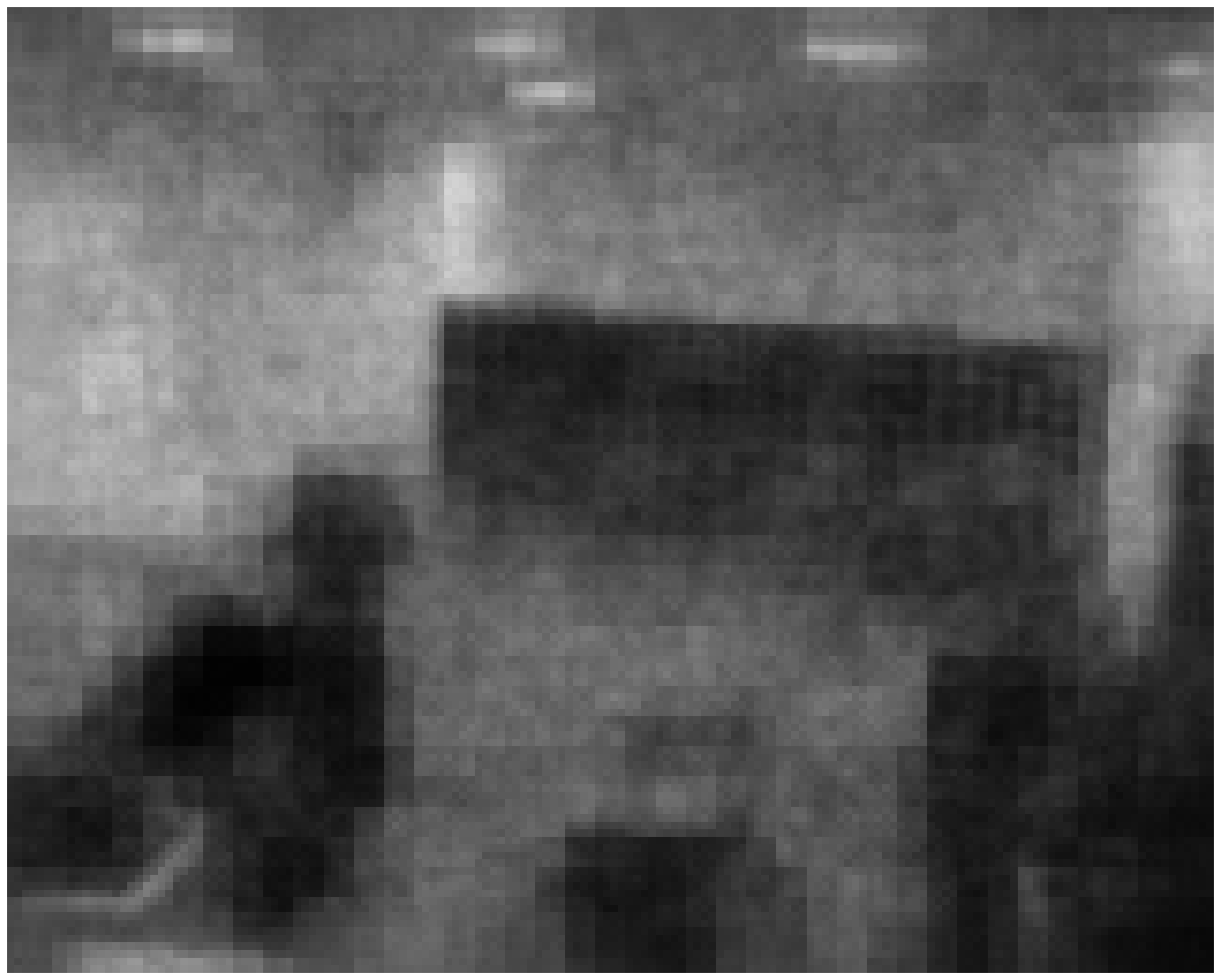}\\
					\hspace*{\fill}\makebox[0pt]{original }\hspace*{\fill}
					\hspace*{\fill}\makebox[0pt]{noisy}\hspace*{\fill}
					\hspace*{\fill}\makebox[0pt]{SLMA}\hspace*{\fill}\\
					\includegraphics[width=25mm]{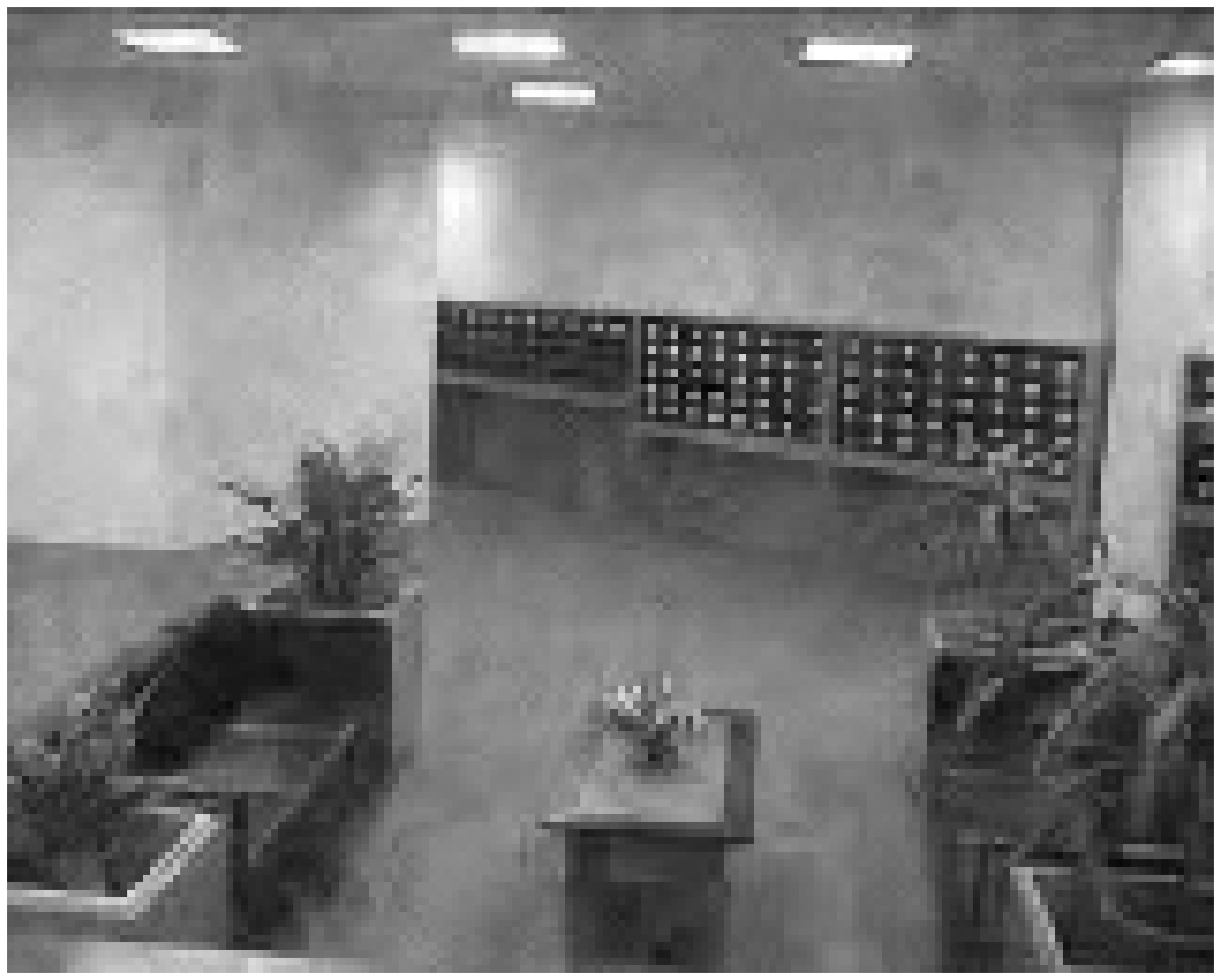}
					\includegraphics[width=25mm]{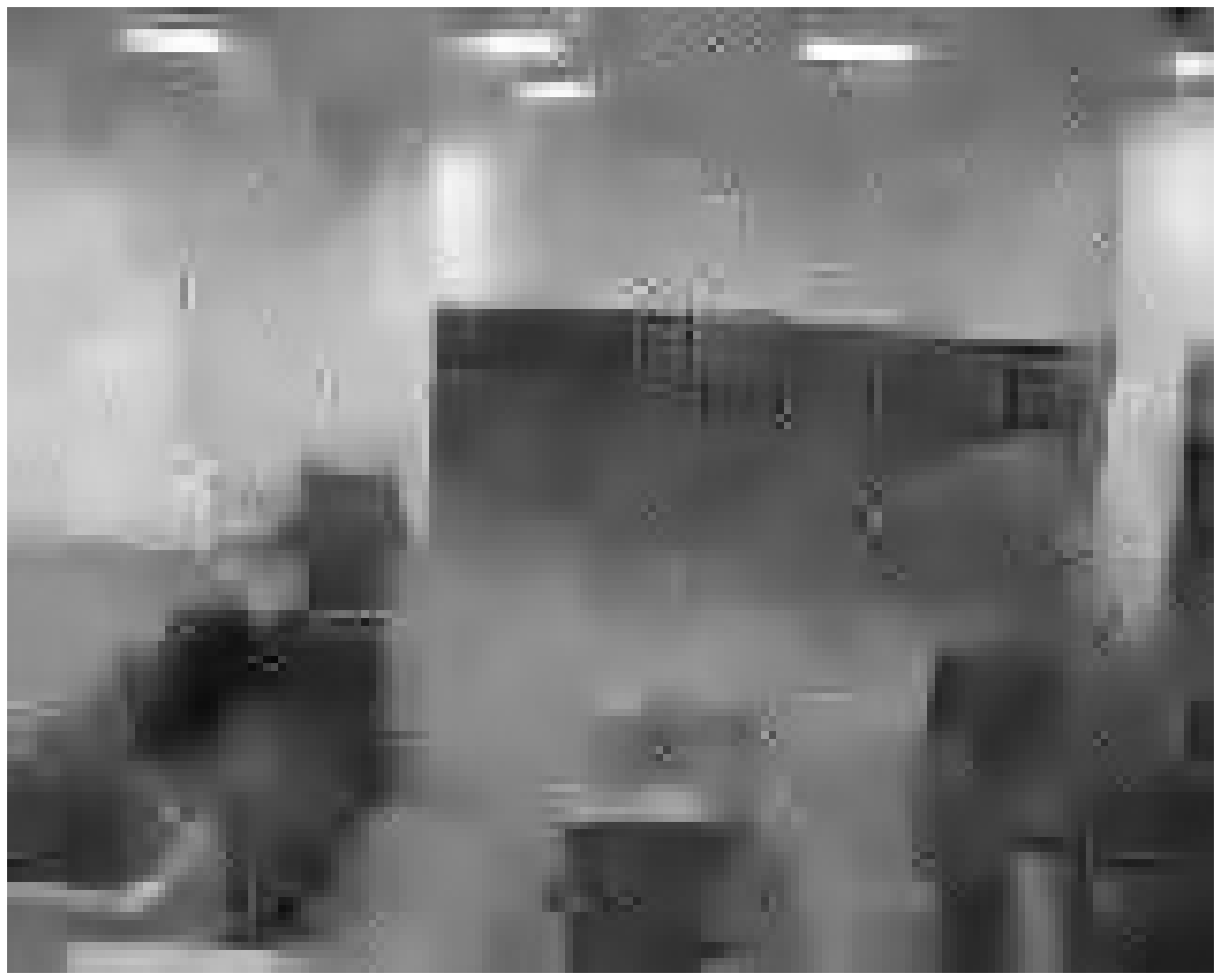}
					\includegraphics[width=25mm]{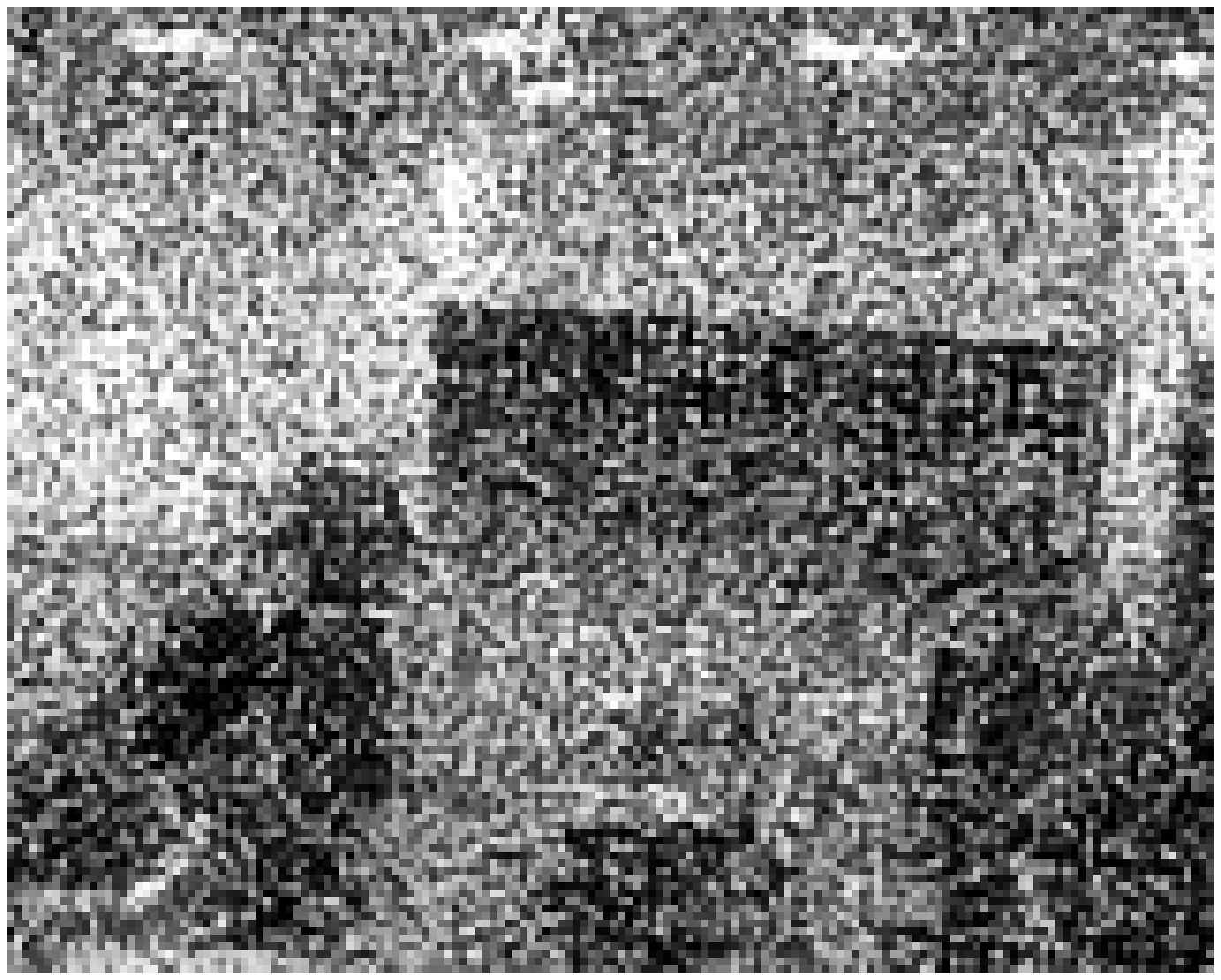}\\
					\hspace*{\fill}\makebox[0pt]{ReLD}\hspace*{\fill}
        			\hspace*{\fill}\makebox[0pt]{VBM3D}\hspace*{\fill}
        			\hspace*{\fill}\makebox[0pt]{MLP}\hspace*{\fill}
						\end{tabular}
					
					\caption{Lobby}\label{VisualLobby}
				\end{subfigure}
				
				\caption{Visual comparison of denoising performance for Curtain and Lobby dataset for very large Gaussian noise ($\sigma=70$)}
				
\end{figure}

\begin{figure*}
	\centering
	\begin{tabular}{cc}
		\includegraphics[width=62mm]{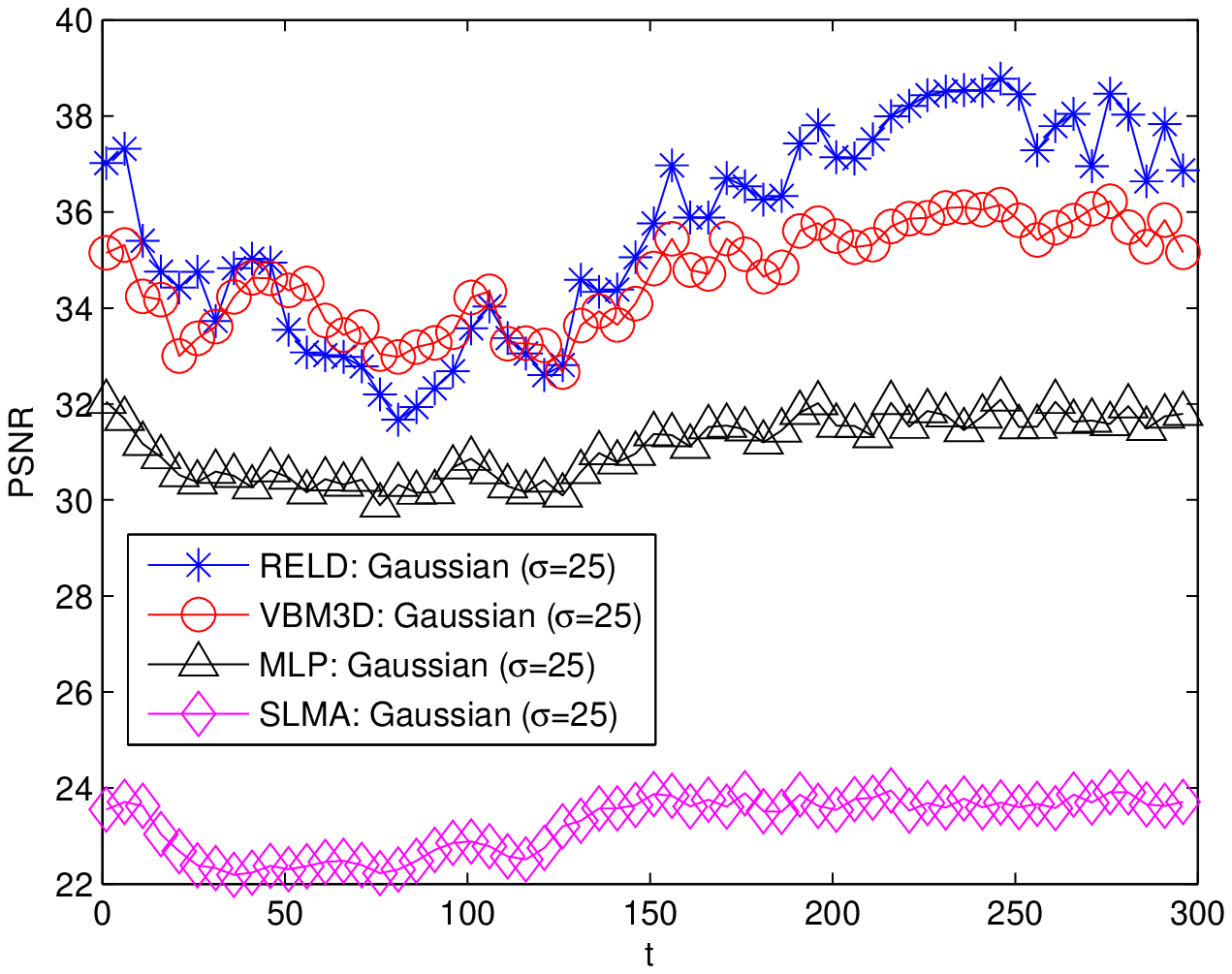}
		\includegraphics[width=62mm]{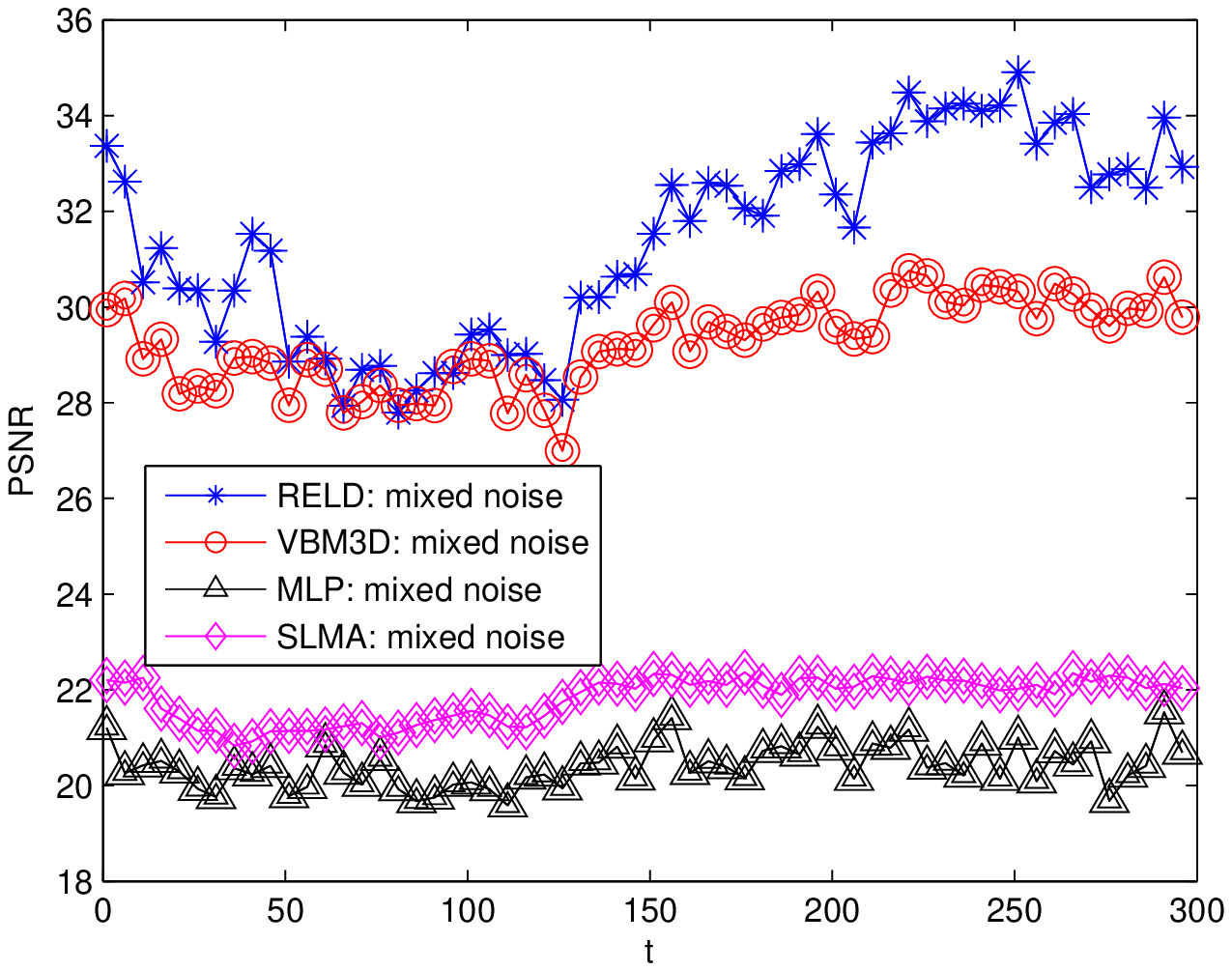}
		\includegraphics[width=62mm]{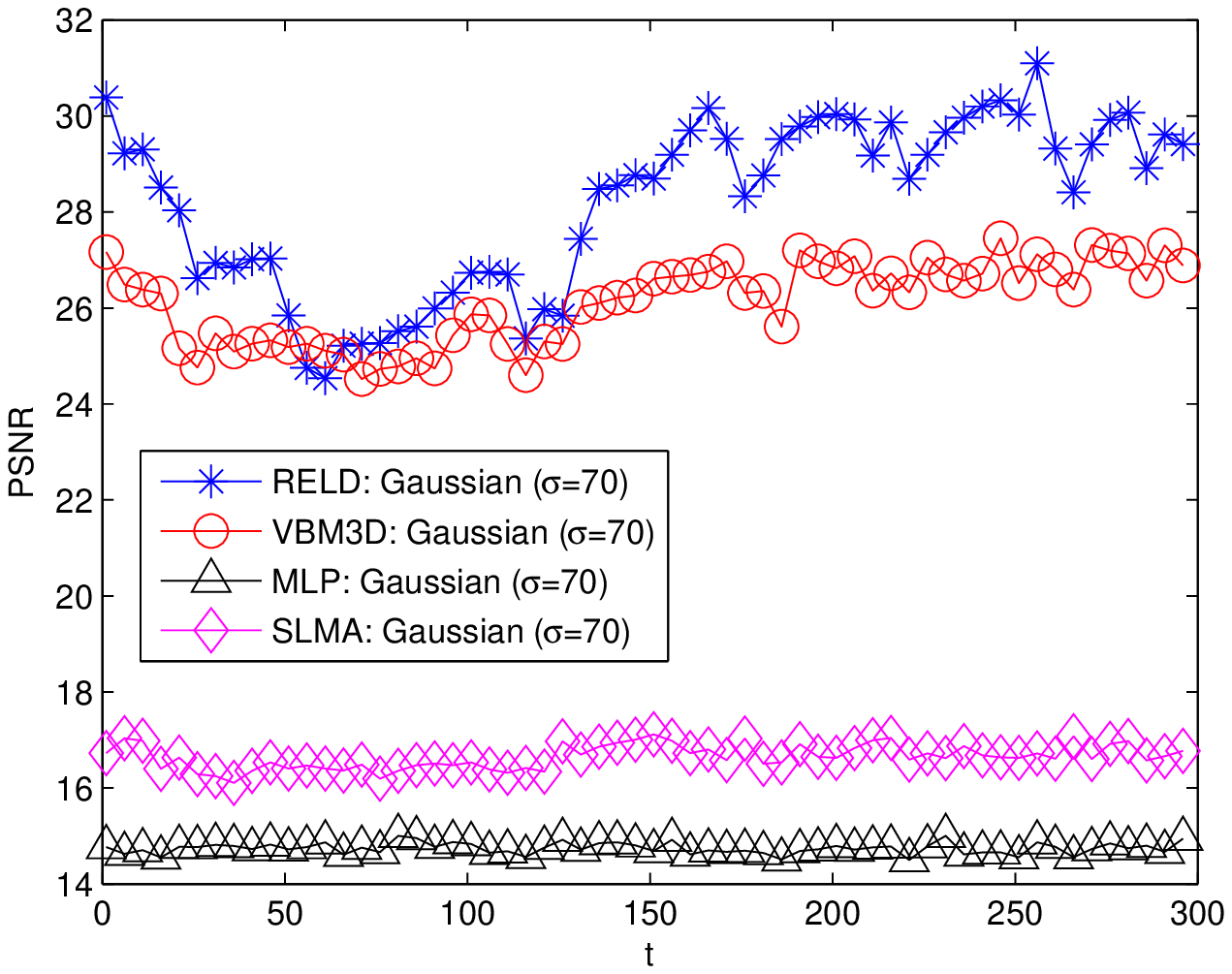}\\	
	    \hspace*{\fill}\makebox[0pt]{(a)}\hspace*{\fill}
		\hspace*{\fill}\makebox[0pt]{(b)}\hspace*{\fill}
		\hspace*{\fill}\makebox[0pt]{(c)}\hspace*{\fill}
	\end{tabular}
\vspace{-0.3cm}
	\caption{Frame-wise PSNR for Curtain dataset with different noise level: (a) Gaussian noise with $\sigma=25$, (b) Gaussian noise ($\sigma=25$) plus salt and pepper noise, (c) Gaussian noise with $\sigma=70$.}\label{CurPSNR}
\vspace{-0.2cm}
\end{figure*}

\begin{figure*}
	\centering
	\begin{tabular}{cc}
		\includegraphics[width=62mm]{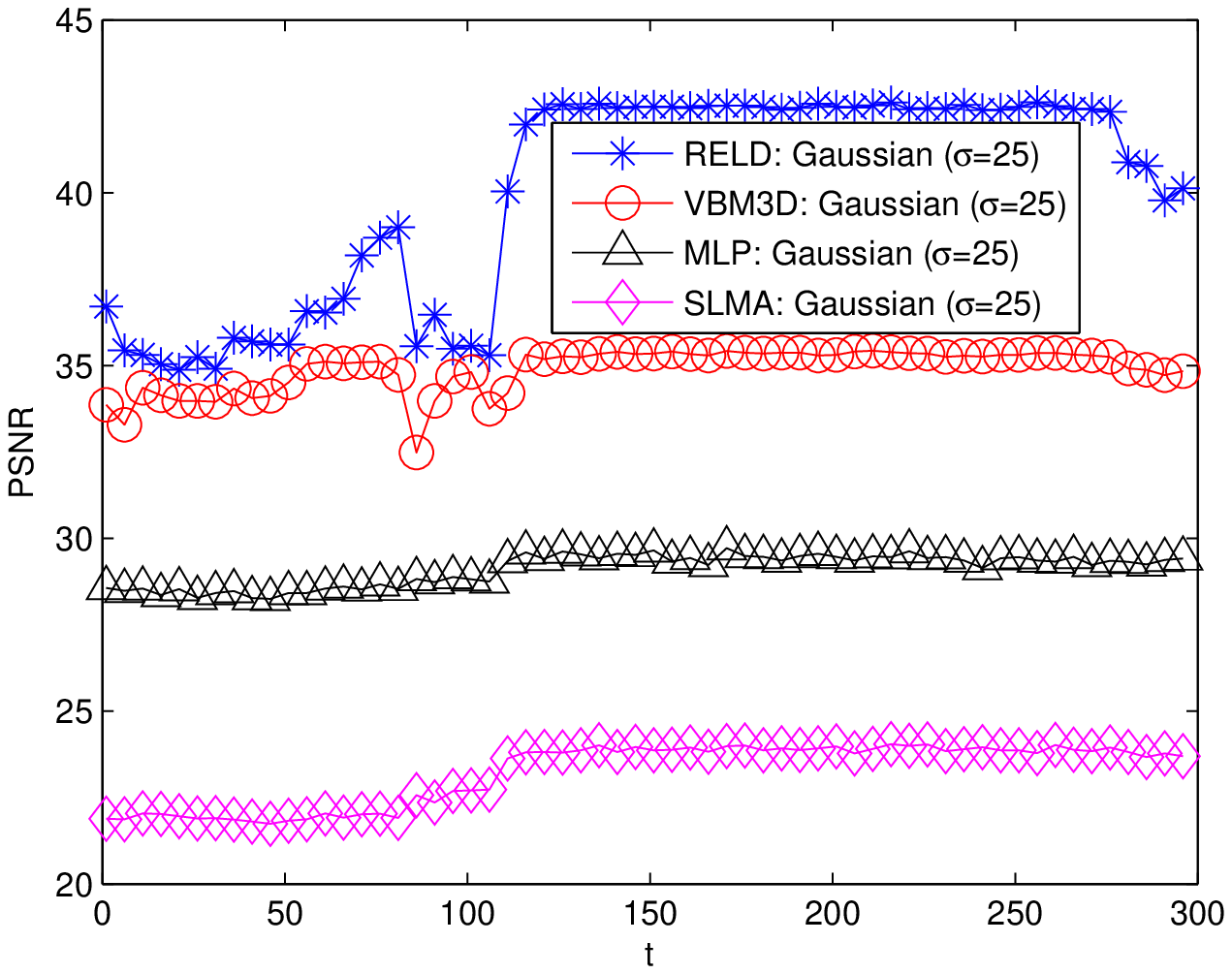}
		\includegraphics[width=62mm]{Lobby25PlusSP.eps}
		\includegraphics[width=62mm]{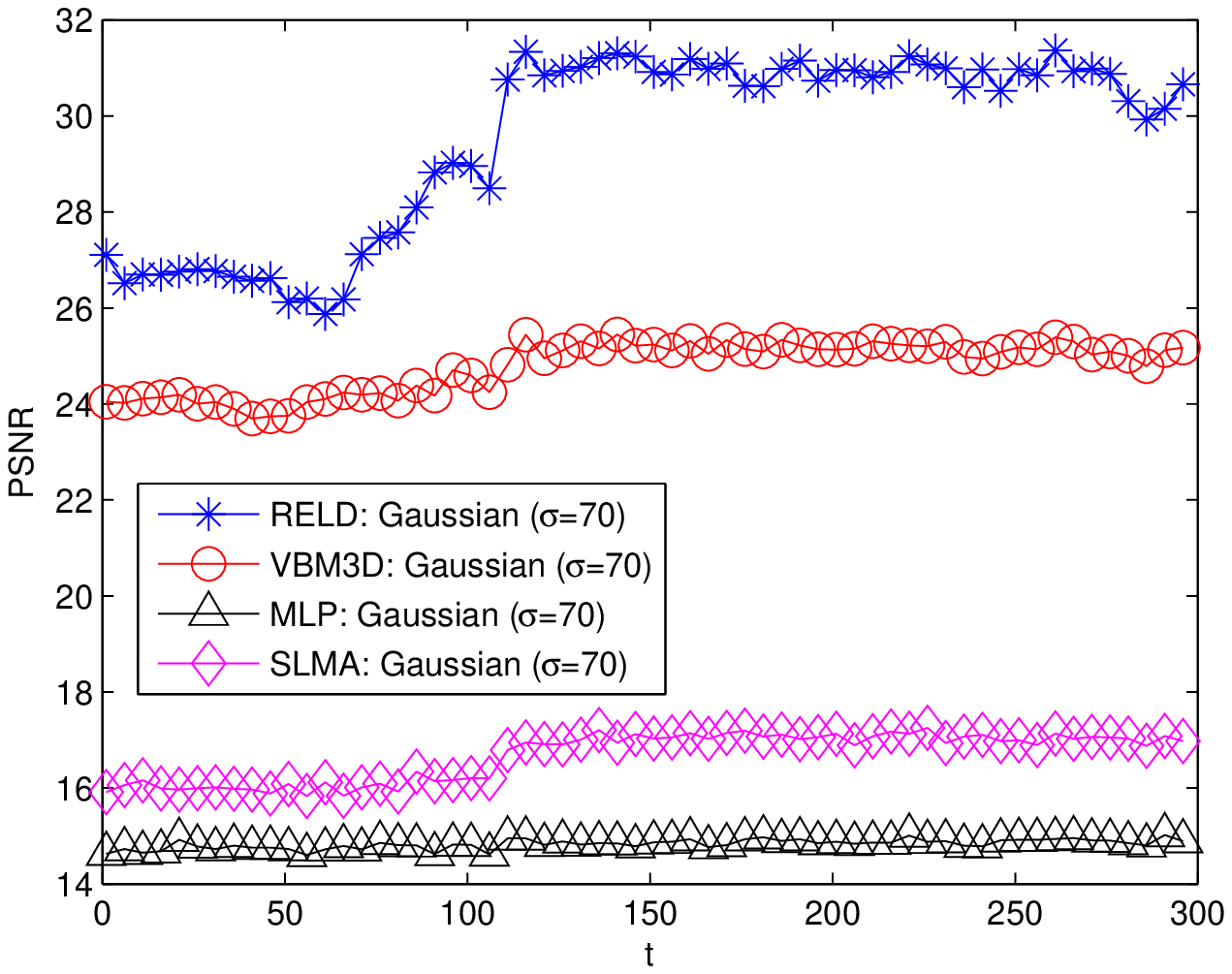}\\
	    \hspace*{\fill}\makebox[0pt]{(a)}\hspace*{\fill}
		\hspace*{\fill}\makebox[0pt]{(b)}\hspace*{\fill}
		\hspace*{\fill}\makebox[0pt]{(c)}\hspace*{\fill}
	\end{tabular}
\vspace{-0.3cm}
	\caption{Frame-wise PSNR for Lobby dataset with different noise level: (a) Gaussian noise with $\sigma=25$, (b) Gaussian noise ($\sigma=25$) plus salt and pepper noise, (c) Gaussian noise with $\sigma=70$.}\label{LobPSNR}
\vspace{-0.2cm}
\end{figure*}

First, we compare the performance of our propsed denoising framework with video layering performed using eiher ReProCS (our proposed algorithm), or using the other robust PCA agorithms - PCP, NCRPCA, and GRASTA. We call the respective algorithms ReLD, PCP-LD, NCRPCA-LD, and GRASTA-LD for short. We test the algorithms on the Waterfall dataset (downloaded from Youtube \url{https://www.youtube.com/watch?v=UwSzu_0h7Bg}). Besides these Laying-Denoing algorithms, we also compare with VBM3D and a neural network image denoising method, Multi Layer Perceptron (MLP). The codes for algorithms being compared are downloaded from the authors' webpages. The available MLP code contains parameters that are trained solely from image patches that were corrupted with Gaussian noise with $\sigma =25$ and hence the denoising performance is best with $\sigma =25$ and deteriorates for other noise levels. 
The video is a background scene without foreground, and hence has no sparse component. We add i.i.d. Gaussian noise with different variance onto the video. Since there is no foreground in the video, the splitting phase can generate a sparse layer which basically consists of the large-magnitude part of the Gaussian noise. The denoising operation followed on such layer does not have the problem of degrading the video quality since this layer is foreground-free.

The video consists of 650 frames and the images are of size $1080\times 1920$. To speed up the algorithms, we first test on the under-sampled data which has image size of $108\times 192$. As can be seen from TABLE \ref{WaterfallSmall}, ReLD has the best denoising performance. We also compare PSNRs using $\hat{\bm{L}}_{\text{denoised}}$ and $\hat{\bm{I}}_{\text{denoised}}$, and we find that using $\hat{\bm{L}}_{\text{denoised}}$ shows an advantage when the noise variance is very large. We then test the algorithms on larger image sizes. To avoid out-of-memory in computation, we only use 100 frames of data. In TABLE \ref{WaterfallMid} and TABLE \ref{WaterfallLarge} we summarize the result for images of size $540 \times 960$ and $1080 \times 1920$ (original), respectively. We notice that, on very large data set (TABLE \ref{WaterfallLarge}), using VBM3D without video-layering algorithms achieves the best denoising performance. This may due to the fact that with larger image size, VBM3D has better chance to find similar image blocks.

Next we thoroughly compare the denoising performance on two more dataset -- curtain and lobby which are available at \url{http://www.ece.iastate.edu/~hanguo/denoise.html}. The algorithms being compared are ReLD, VBM3D, MLP and SLMA. The noise being added to the original image frames are Gaussian ($\sigma=25$), Gaussian ($\sigma=25$) plus salt and pepper noisy, and Gaussian ($\sigma=70$). 
The input $\sigma$ for all algorithms is estimated from the noisy data rather then given the true value. We compute the frame-wise PSNRs (using $\hat{\bm{I}}_{\text{denoised}}$) for each case in Fig. \ref{CurPSNR} and Fig. \ref{LobPSNR} and show sample visual comparisons in Fig. \ref{VisualCurtain} and Fig. \ref{VisualLobby}. We can see in Fig. \ref{CurPSNR} and Fig. \ref{LobPSNR} that ReLD outperforms all other algorithms in all three noise level -- the PSNR is the highest in almost all image frames. Visually, ReLD is able to recover more details of the images while other algorithms either fail or cause severe blurring effect. We test the algorithms on two more datasets (the fountain dataset and escalator dataset). The noise being added to the original video is Gaussian, with standard deviation $\sigma$ increases from $25$ to $70$. We present the PSNRs in Table \ref{TabPSNR}. 
\end{document}